\documentclass[11pt, a4paper, logo, copyright, nonumbering]{mll_style}
\usepackage{color}
\usepackage{epsfig}
\usepackage{graphicx}

\usepackage{booktabs}  %
\usepackage{tabularx}               %
\usepackage{ltablex}
\usepackage{tabularray}
\newcolumntype{C}{>{\centering\arraybackslash}X}
\usepackage{multirow}               %
\usepackage{diagbox}                %
\usepackage{hhline}                 %
\usepackage{color}                  %

\usepackage{booktabs}
\usepackage{extarrows}
\usepackage{makecell}
\usepackage{wrapfig}
\usepackage{colortbl}
\usepackage{longtable}
\usepackage{tabularray}
\usepackage[most]{tcolorbox}
\tcbuselibrary{skins} %
\usepackage{tcolorbox}

\usepackage{amssymb}

\usepackage{adjustbox}
\usepackage{array}
\usepackage{floatflt}

\usepackage{bm}
\usepackage{nicefrac}
\usepackage{microtype}

\usepackage{changepage}
\usepackage{extramarks}
\usepackage{fancyhdr}
\usepackage{lastpage}
\usepackage{setspace}
\usepackage{soul}
\usepackage{xspace}

\usepackage{url}

\usepackage{enumerate}
\usepackage{enumitem}  %

\usepackage{makecell}

\usepackage{pifont} %

\usepackage{algorithm}
\usepackage{algpseudocode}
\usepackage{xcolor}

\usepackage[symbol]{footmisc}

\usepackage{caption}
\usepackage{subcaption}

\usepackage{scalefnt}

\usepackage{fontawesome5}

\usepackage{siunitx}

\usepackage{listings}

\newcolumntype{L}[1]{>{\raggedright\let\newline\\\arraybackslash\hspace{0pt}}m{#1}}
\newcolumntype{R}[1]{>{\raggedleft\let\newline\\\arraybackslash\hspace{0pt}}m{#1}}

\newcommand{\ignore}[1]{}

\makeatletter
\DeclareRobustCommand\onedot{\futurelet\@let@token\@onedot}
\def\@onedot{\ifx\@let@token.\else.\null\fi\xspace}

\makeatother

\definecolor{MyBlue}{rgb}{0.46, 0.50, 0.61}
\definecolor{MyDarkBlue}{rgb}{0,0.08,0.8}
\definecolor{MyDarkGreen}{RGB}{45,155,45}
\definecolor{MyDarkRed}{rgb}{0.8,0.02,0.02}
\definecolor{MyOrange}{rgb}{1.0, 0.4, 0.2}
\definecolor{MyPurple}{RGB}{111,0,255}
\definecolor{MyRed}{rgb}{0.8,0.0,0.0}
\definecolor{MyGold}{rgb}{0.75,0.6,0.12}
\definecolor{MyDarkgray}{rgb}{0.66, 0.66, 0.66}
\definecolor{MyBrown}{rgb}{0.65, 0.16, 0.16}
\definecolor{MyMutedRose}{rgb}{0.58, 0.29, 0.35}
\definecolor{JiayuanColor}{rgb}{0.60,0.43,0.48}
\definecolor{erranColor}{rgb}{24, 40, 113}

\definecolor{citecolor}{HTML}{696FAD}

\newif\ifpropositionfirstitem
\propositionfirstitemtrue

\newcommand{\cmark}{\ding{51}}%
\newcommand{\xmark}{\ding{55}}%

\definecolor{bggray}{HTML}{F5F5F5}
\definecolor{pvdblue}{HTML}{DAE8FC}
\definecolor{RoseQuartzBg}{HTML}{F7CAC9}
\definecolor{RoseQuartz}{HTML}{F5A798}
\definecolor{Serenity}{HTML}{92A8D1}
\definecolor{OrangeRed}{rgb}{1.0, 0.27, 0.0}
\definecolor{RoyalBlue}{cmyk}{1, 0.50, 0, 0}
\definecolor{Turquoise}{HTML}{0F4C81}
\definecolor{mint}{rgb}{0.24, 0.71, 0.54}
\definecolor{green}{rgb}{0.0, 0.120, 0.0}

\newdimen\abovecrulesep
\newdimen\belowcrulesep
\makeatletter
\patchcmd{\@@@cmidrule}{\aboverulesep}{\abovecrulesep}{}{}
\patchcmd{\@xcmidrule}{\belowrulesep}{\belowcrulesep}{}{}
\makeatother

\definecolor{mybluetitle}{HTML}{4B527E} %

\definecolor{mygreen}{RGB}{0,150,0}
\definecolor{boxbackground}{HTML}{F0F7FF}  %
\definecolor{boxborder}{HTML}{D0D9E5}      %
\definecolor{accentblue}{HTML}{4A86E8}     %
\definecolor{lightblue}{HTML}{EEF3FF}  %
\definecolor{bordergray}{HTML}{CCCCCC}  %
\definecolor{headerblue}{HTML}{2C5AA0}  %

\definecolor{lavenderframe}{HTML}{E6E6FA}  %
\definecolor{lighterlav}{HTML}{F5F5FF}  %
\definecolor{codegray}{rgb}{0.5,0.5,0.5}  %
\definecolor{codepurple}{HTML}{483D8B}  %
\definecolor{backcolour}{HTML}{F5F5FF}  %
\lstdefinestyle{mystyle}{
    backgroundcolor=\color{backcolour},
    commentstyle=\color{headerblue},
    keywordstyle=\color{codepurple},
    numberstyle=\tiny\color{codegray},
    stringstyle=\color{codepurple},
    basicstyle=\ttfamily\scriptsize,
    breakatwhitespace=false,
    breaklines=true,
    captionpos=b,
    keepspaces=true,
    frame=none,
    numbersep=5pt,
    showspaces=false,
    showstringspaces=false,
    showtabs=false,
    tabsize=2
}

\definecolor{jsonkey}{RGB}{44, 130, 201}     %
\definecolor{jsonstring}{RGB}{255, 140, 0}   %
\definecolor{jsonnumber}{RGB}{34, 139, 34}   %

\lstdefinelanguage{json}{
    basicstyle=\ttfamily\small,
    numbers=left,
    numberstyle=\tiny\color{gray},
    stepnumber=1,
    numbersep=5pt,
    showstringspaces=false,
    breaklines=true,
    frame=none,
    backgroundcolor=\color{gray!5},
    literate=
     *{:}{{{\color{jsonkey}:}}}{1}
      {,}{{{\color{jsonkey},}}}{1}
      {"}{{{\color{jsonstring}"}}}{1}
      {[}{{{\color{jsonkey}[}}}{1}
      {]}{{{\color{jsonkey}]}}}{1}
      {0}{{{\color{jsonnumber}0}}}{1}
      {1}{{{\color{jsonnumber}1}}}{1}
      {2}{{{\color{jsonnumber}2}}}{1}
      {3}{{{\color{jsonnumber}3}}}{1}
      {4}{{{\color{jsonnumber}4}}}{1}
      {5}{{{\color{jsonnumber}5}}}{1}
      {6}{{{\color{jsonnumber}6}}}{1}
      {7}{{{\color{jsonnumber}7}}}{1}
      {8}{{{\color{jsonnumber}8}}}{1}
      {9}{{{\color{jsonnumber}9}}}{1}
}

\newtcblisting{jsonbox}{
  listing engine=listings,
  colback=gray!3!white,
  colframe=gray!75!black,
  boxrule=0.4mm,
  arc=2mm,
  outer arc=2mm,
  breakable,
  enhanced,
  listing only,
  listing options={language=json}
}

\newtcolorbox{promptbox}[2][]{ %
    enhanced,
    breakable,
    boxsep=5pt,
    left=9pt,
    right=7pt,
    top=5pt,
    bottom=5pt,
    colback=boxbackground,
    colframe=boxborder,
    boxrule=0.5pt,
    arc=4pt,
    frame hidden, %
    borderline west={3pt}{0pt}{accentblue},
    shadow={0.5pt}{0.5pt}{1.5pt}{black!10},
    fontupper=\normalsize,
    title=#2, %
    colbacktitle=accentblue, %
    coltitle=white,         %
    fonttitle={\fontsize{9}{11}\selectfont\bfseries}, %
    attach boxed title to top left={yshift=-2.5mm, xshift=3.2mm},
    boxed title style={
        enhanced,
        left=3pt,
        right=3pt,
        top=1pt,    %
        bottom=1pt, %
        boxsep=2pt,
        arc=3pt,
        boxrule=0pt,
        colback=accentblue,
    },
    #1 %
}

\newtcolorbox{notitlepromptbox}[1][]{
    enhanced,
    breakable,
    boxsep=5pt,          %
    left=9pt,            %
    right=7pt,           %
    top=5pt,             %
    bottom=5pt,          %
    colback=boxbackground,
    colframe=boxborder,
    boxrule=0.5pt,
    arc=4pt,             %
    frame hidden,
    borderline west={3pt}{0pt}{accentblue},  %
    shadow={0.5pt}{0.5pt}{1.5pt}{black!10},  %
    notitle,
    fontupper=\normalsize,    %
    #1
}

\newtcolorbox{onebox}[2][]{
    enhanced, 
    center title,
    left*=0pt, right*=0pt,
    boxsep=2pt, left=5pt, right=5pt,
    skin first=enhanced,
    skin middle=enhanced,
    skin last=enhanced,
    colframe = mybluetitle!90,
  colback  = mybluetitle!10,
    fonttitle=\bfseries\rmfamily\fontfamily{phv}\selectfont,
    title={\footnotesize\strut{#2}  \refstepcounter{subsubsection} \addcontentsline{toc}{subsubsection}{\string\numberline{\thesubsubsection}#2}
    },
    #1
    }

\usepackage{algorithm}          
\usepackage{algpseudocode}      
\usepackage{amsmath}            
\usepackage{amsfonts}           
\usepackage{bm}   

\usepackage{amsfonts}
\usepackage[numbers]{natbib}
\usepackage{graphicx}
\usepackage{xspace}
\usepackage{xcolor}
\definecolor{highlightgray}{RGB}{220, 220, 220}
\usepackage{adjustbox}
\usepackage{array}
\usepackage{float}
\usepackage{geometry}
\usepackage{dblfloatfix}
\usepackage{enumitem}
\usepackage{setspace}
\usepackage{booktabs}
\usepackage{multirow}
\usepackage{longtable}
\usepackage{tabularx}
\usepackage{xltabular}
\usepackage{threeparttable}
\usepackage{siunitx}
\usepackage{amsmath,array}
\newcommand{\NA}{--}

\usepackage{pdflscape}

\usepackage{amsmath}
\usepackage{amssymb}
\usepackage{amsfonts}
\usepackage{mathtools}
\usepackage{bm}
\usepackage{dsfont}

\usepackage{subcaption}
\usepackage{tikz}
\usepackage{pgfplots}
\pgfplotsset{compat=1.18}
\usepackage{algorithm}
\usepackage{algpseudocode}
\usepackage{listings}

\usepackage{cleveref}
\hypersetup{colorlinks=true}

\usepackage{enumitem}
\usepackage{todonotes}
\usepackage{fontawesome5}
\usepackage[most]{tcolorbox}
\usepackage{CJKutf8}
\usepackage{lipsum}

\algdef{SE}[FORALL]{ForAll}{EndForAll}[1]{\algorithmicforall\ #1\ }{}
\algtext*{EndForAll}

\definecolor{tableblue}{RGB}{201,226,239}

\makeatletter
\def\@BTrule[#1]{%
  \ifx\longtable\undefined
    \let\@BTswitch\@BTnormal
  \else\ifx\hline\LT@hline
    \nobreak
    \let\@BTswitch\@BLTrule
  \else
     \let\@BTswitch\@BTnormal
  \fi\fi
  \global\@thisrulewidth=#1\relax
  \ifnum\@thisruleclass=\tw@\vskip\@aboverulesep\else
  \ifnum\@lastruleclass=\z@\vskip\@aboverulesep\else
  \ifnum\@lastruleclass=\@ne\vskip\doublerulesep\fi\fi\fi
  \@BTswitch}
\makeatother

\addto\extrasenglish{
}

 {\begin{list}{}%
         {\setlength{\leftmargin}{#1}}%
         \item[]%
 }
 {\end{list}}

\usepackage[toc,page,header]{appendix}
\usepackage{minitoc}

\renewcommand \partname{}

\reportnumber{001} %

\title{\centering Holi-Spatial: Evolving Video Streams into Holistic 3D Spatial~Intelligence
}

\date{}

\author[*]{
Yuanyuan Gao$^{1, 2*}$, Hao Li$^{2, 5*}$, Yifei Liu$^{1, 6*}$, Xinhao Ji$^{1, 4*}$, Yuning Gong$^{1, 7*}$, Yuanjun Liao$^{7}$, Fangfu Liu$^{8}$, Manyuan Zhang$^{9}$, Yuchen Yang$^{10}$, Dan Xu$^{11}$, Xue Yang$^{3}$, Huaxi Huang$^{1}$, Hongjie~Zhang$^{1}$, Ziwei Liu$^{5}$, Xiao Sun$^{1}$, Dingwen Zhang$^{2,\dag}$, Zhihang Zhong$^{3,\dag}$

{\small $^*$Equal Contribution; $^{\dag}$Corresponding Authors}
\\
\small $^1$Shanghai~AI~Lab,  
$^2$Northwestern~Polytechnical~University,  
$^3$Shanghai~Jiao~Tong~University,  
$^4$Peking~University,  
$^5$Nanyang~Technological~University,  
$^6$Beihang~University,  
$^7$Sichuan~University,  
$^8$Tsinghua~University, 
$^9$The~Chinese~University~of~Hong~Kong, 
$^{10}$Fudan~University, 
$^{11}$Hong~Kong~University~of~Science~and~Technology 
\\
\vspace{-10pt}
}

\begin{abstract}
The pursuit of spatial intelligence fundamentally relies on access to large-scale, fine-grained 3D data. However, existing approaches predominantly construct spatial understanding benchmarks by generating question–answer (QA) pairs from a limited number of manually annotated datasets, rather than systematically annotating new large-scale 3D scenes from raw web data. As a result, their scalability is severely constrained, and model performance is further hindered by domain gaps inherent in these narrowly curated datasets.
In this work, we propose \textbf{Holi-Spatial}, the first fully automated, large-scale, spatially-aware multimodal dataset, constructed from raw video inputs without human intervention, using the proposed data curation pipeline. Holi-Spatial supports multi-level spatial supervision, ranging from geometrically accurate 3D Gaussian Splatting (3DGS) reconstructions with rendered depth maps to object-level and relational semantic annotations, together with corresponding spatial Question–Answer (QA) pairs.
Following a principled and systematic pipeline, we further construct \textbf{Holi-Spatial-4M}, the first large-scale, high-quality 3D semantic dataset, containing 12K optimized 3DGS scenes, 1.3M 2D masks, 320K 3D bounding boxes, 320K instance captions, 1.2M 3D grounding instances, and 1.2M spatial QA pairs spanning diverse geometric, relational, and semantic reasoning tasks.
Holi-Spatial demonstrates exceptional performance in data curation quality, significantly outperforming existing feed-forward and per-scene optimized methods on datasets such as ScanNet, ScanNet++, and DL3DV. Furthermore, fine-tuning Vision-Language Models (VLMs) on spatial reasoning tasks using this dataset has also led to substantial improvements in model performance.

\vspace{8pt}

\textbf{Website}: \href{https://visionary-laboratory.github.io/holi-spatial/}{https://visionary-laboratory.github.io/holi-spatial/}\\~
\textbf{Code}: \href{https://github.com/Visionary-Laboratory/Holi-Spatial}{https://github.com/Visionary-Laboratory/Holi-Spatial}\\~
\textbf{Email: } \texttt{zhongzhihang95@gmail.com}

\end{abstract}

\begin{document}
\begin{CJK*}{UTF8}{gbsn}

\doparttoc %
\faketableofcontents %

\maketitle

\begin{center}
    \centering
    \includegraphics[width=1\textwidth]{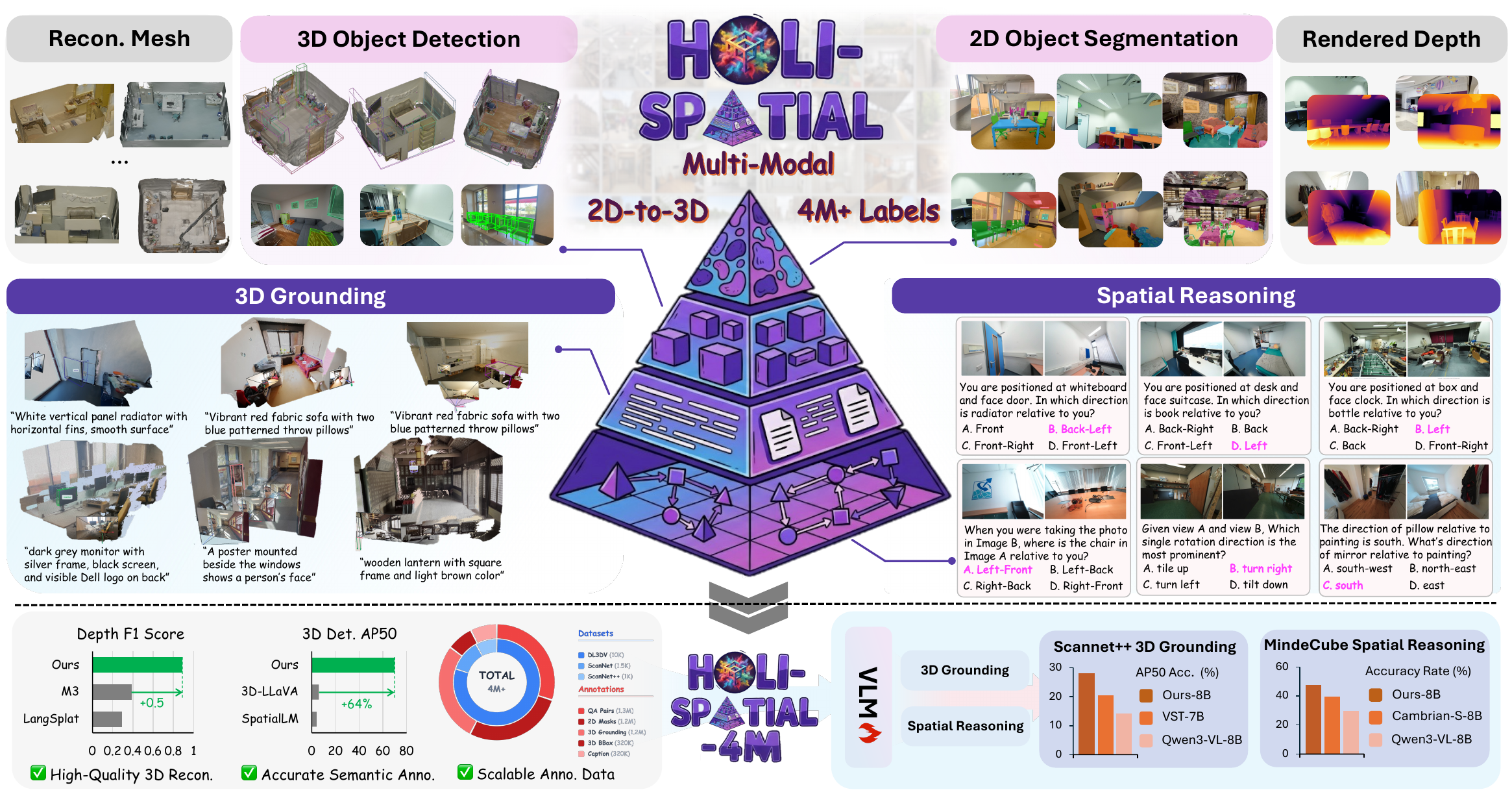} 
    \vspace{-0.1in}
    \captionof{figure}{\textbf{We introduce \textbf{Holi-Spatial}, the first fully automated pipeline capable of converting raw video streams into holistic 3D spatial annotations without human intervention.} Compared to state-of-the-art methods, Holi-Spatial achieves a significant leap in annotation quality, improving multi-view depth estimation by 0.5 F1 and boosting 3D detection AP${50}$ by a remarkable 64\% on ScanNet~\cite{dai2017scannet}. Based on this, we introduce \textbf{Holi-Spatial-4M}, a large-scale dataset that effectively empowers Vision-Language Models. As shown, fine-tuning Qwen3-VL on Holi-Spatial-4M leads to state-of-the-art performance, with a 15\% AP${50}$ gain on ScanNet++~\cite{yeshwanth2023scannet++} and a 7.9\% accuracy rise on MMSI-Bench~\cite{yang2025mmsi}. \textbf{Importantly, because the entire annotation pipeline is automatic, it can be further scaled up as resources permit.}}
    \label{fig:teaser}
\end{center}

\section{Introduction}
Spatial intelligence~\cite{feng2025survey} is a fundamental bridge toward enabling large models to understand the real 3D world. It requires large multimodal models (LMMs)~\cite{achiam2023gpt, Qwen3-VL,team2023gemini,Qwen2.5-VL} to move beyond 2D, language-centric perception and develop robust 3D spatial abilities to perceive, ground, and reason about the 3D world from visual inputs. Such capabilities hold great promise for a wide range of real-world applications, including robotic manipulation~\cite{zhang2025spatial,qu2025spatialvla} and navigation~\cite{he2025strider}, scene editing~\cite{wang2025embodiedgen}, and augmented reality~\cite{jiang2025anysplat}.

However, a key limitation is the scarcity and imbalance of raw spatial data. Prior methods~\cite{wu2025spatial,cai2025scaling,yin2025spatial,yang2025visual} typically curate spatial supervision by generating QA pairs from a small set of manually annotated 3D datasets (\textit{e.g.}, ScanNet~\cite{dai2017scannet} and ScanNet++~\cite{yeshwanth2023scannet++}) or by naively applying feed-forward perception models~\cite{carion2025sam} to single-image data~\cite{gupta2019lvis}. While these strategies improve over general VLMs~\cite{Qwen3-VL,Qwen2.5-VL,team2023gemini}, they are difficult to scale due to reliance on specialized scanning hardware and human-in-the-loop annotation, and they often provide limited semantic coverage (\textit{e.g.}, only 50 labeled classes in ScanNet~\cite{dai2017scannet}).

To address these limitations, we note that recent advances in relevant AI tools~\cite{ravi2024sam2, carion2025sam,team2023gemini,wang2025vggt, lin2025depth, keetha2026mapanything,li2025iggt} have exceeded expectations; by systematically composing them, we may build an automated spatial annotation engine that can even outperform human annotations, enabling a positive data flywheel. Thus, we present \textbf{Holi-Spatial}, a fully automated framework that converts raw video streams into high-fidelity 3D geometry together with holistic semantic annotations \emph{without} requiring any explicit 3D sensors or human-in-the-loop labeling. As summarized in~\cref{tab:capability_comparison}, Holi-Spatial unifies a broad set of spatial tasks, including 3D reconstruction, novel view synthesis (NVS), depth rendering, 2D instance segmentation, instance captioning, 3D bounding boxes, 3D grounding, and spatial QA.

Holi-Spatial is composed of three stages. (i) \textit{Geometric Optimization:} we initialize from monocular priors (Depth-Anything-V3~\cite{lin2025depth}) and optimize a 3D Gaussian Splatting (3DGS) scene under geometric supervision to sharpen structure and suppress floaters. (ii) \textit{Image-level Perception:} we sample keyframes, use a VLM to infer open-vocabulary categories, which guides SAM3 to produce high-quality open-set masks per-image. (iii) \textit{Scene-level Lift and Refinement:} We lift the 2D masks into 3D by back-projecting their pixels using rendered depth and camera intrinsics, and transforming the resulting points into the world frame using recovered camera poses. The resulting 3D points serve as instance candidates. We merge redundant candidates across views by checking their bounding box IoU, and use a VLM-based agent to filter low-confidence candidates; for the merged and most reliable instances, we generate detailed captions and further construct grounding and QA pairs for training VLM's 3D Grounding and Spatial Reasoning ability.
\begin{wrapfigure}{r}{0.6\textwidth}
  \centering
  \includegraphics[width=0.58\textwidth]{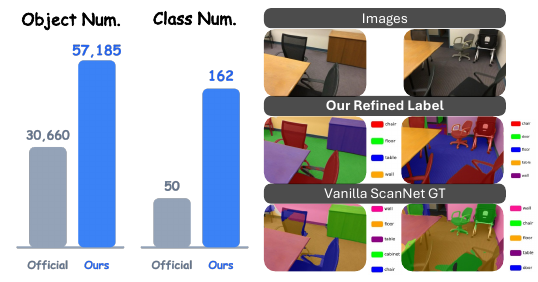}
  \caption{\textbf{Comparison of our refined annotations with the official annotations on the ScanNet dataset~\cite{dai2017scannet}.} Our method achieves more accurate and sharper segmentation masks, as well as improved category recognition.}
  \label{fig:compare_scannet_official}
  \vspace{-0.1in}
\end{wrapfigure}

We evaluate the proposed curation pipeline on established benchmarks, including ScanNet~\cite{dai2017scannet}, ScanNet++~\cite{yeshwanth2023scannet++}, and DL3DV-10K~\cite{ling2024dl3dv}, where we create open-vocabulary GT by additionally re-annotating the scenes with human supervision to enable reliable evaluation of fine-grained objects. As shown in~\cref{fig:teaser}, Holi-Spatial consistently outperforms both model-based baselines and 3DGS-based optimization approaches; on ScanNet++~\cite{yeshwanth2023scannet++}, we improve multi-view depth estimation by up to 0.5 F1 and boost 3D detection AP$_{50}$ by 64\%.

Leveraging this pipeline, we construct and release \textbf{Holi-Spatial-4M} from diverse video sources (ScanNet~\cite{dai2017scannet}, ScanNet++~\cite{yeshwanth2023scannet++}, and DL3DV-10K~\cite{ling2024dl3dv}). The dataset contains 12K optimized 3DGS scenes with multi-view consistent depth renderings, as well as 1.2M 2D masks, 1.2M 3D Grounding, 320K 3D bounding boxes, 320K instance captions, and 1.3M spatial QA pairs. Beyond scale, our annotations exhibit higher granularity and improved boundaries compared to official labels on Scannet~\cite{dai2017scannet} (~\cref{fig:compare_scannet_official}).
To verify that Holi-Spatial-4M improves spatial intelligence, we fine-tune the Qwen3-VL family~\cite{Qwen3-VL,Qwen2.5-VL} on our curated data. This yields consistent gains on public benchmarks, including a 15\% AP$_{50}$ improvement on ScanNet++~\cite{yeshwanth2023scannet++} for 3D grounding and a 7.9\% accuracy increase on MMSI-Bench~\cite{yang2025mmsi}.

\begin{table}[htbp]
    \centering

    \caption{\textbf{Overview of Pipeline Capabilities.} We compare input modalities and output tasks across different paradigms, ranging from 2D VLMs and 3D-VLMs to 3D-GS-based understanding methods. Holi-Spatial serves as a unified framework supporting diverse spatial tasks without relying on 3D priors.}
    \vspace{0.1cm}
    
    \setlength{\tabcolsep}{4pt} 

    \begin{tabular}{l cc ccccc}
        \toprule[1.5pt]
        \multirow{2}{*}{\textbf{Method}} & 
        \multicolumn{2}{c}{\textbf{Inputs}} & 
        \multicolumn{5}{c}{\textbf{Outputs}} \\
        \cmidrule(lr){2-3} \cmidrule(lr){4-8}
        & \textbf{Images} & \textbf{Point Clouds} & \textbf{Depth} & \textbf{2D Seg.} & \textbf{3D Det.} & \textbf{Grounding} & \textbf{Spatial QA} \\
        \midrule
        
        \multicolumn{8}{l}{\textit{\textbf{2D-VLM Methods}}} \\
        \hspace{2mm}SAM3       & \cmark & \xmark & \xmark & \cmark & \xmark & \xmark & \xmark \\
        \hspace{2mm}SA2VA      & \cmark & \xmark & \xmark & \cmark & \xmark & \cmark & \cmark \\
        \addlinespace[3pt]
        
        \multicolumn{8}{l}{\textit{\textbf{3D-VLM Methods}}} \\
        \hspace{2mm}SpatialLM  & \xmark & \cmark & \xmark & \xmark & \cmark & \cmark & \cmark \\
        \hspace{2mm}LLaVA-3D   & \xmark & \cmark & \xmark & \xmark & \cmark & \cmark & \cmark \\
        \hspace{2mm}SceneScript& \xmark & \cmark & \xmark & \xmark & \cmark & \xmark & \xmark \\
        \addlinespace[3pt]
        
        \multicolumn{8}{l}{\textit{\textbf{3D-GS based Understanding Methods}}} \\
        \hspace{2mm}M3-Spatial & \cmark & \xmark & \cmark & \cmark & \xmark & \cmark & \xmark \\
        \hspace{2mm}LangSplat  & \cmark & \xmark & \cmark & \cmark & \xmark & \xmark & \xmark \\
        \addlinespace[3pt]
        
        \midrule
        \rowcolor{highlightgray}
        \textbf{Holi-Spatial} & \cmark & \cmark & \cmark & \cmark & \cmark & \cmark & \cmark \\
        \bottomrule[1.5pt]
    \end{tabular}
    \label{tab:capability_comparison}
\end{table}

\section{Related Work}

\subsection{Data Scalability in Spatial Intelligence}
Recent advancements in Large Multimodal Models (LMMs) have demonstrated remarkable proficiency in 2D visual understanding and planar reasoning~\cite{team2023gemini,  Qwen3-VL, Qwen2.5-VL}, but still exhibit a significant lag behind human capabilities in Spatial Intelligence~\cite{cai2025holistic,yang2025mmsi,yin2025spatial}.
We attribute this gap primarily to a critical disparity in \textbf{raw data diversity}.
Unlike 2D datasets such as LAION-5B~\cite{schuhmann2022laion} provide \textit{billion-scale} unique image inputs, the spatial intelligence domain suffers from severe scene scarcity.
Although recent spatial datasets such as SenseNova-SI-800K~\cite{cai2025scaling} and VST-4M~\cite{yang2025visual} boast million-level annotations, these are predominantly derived from a tiny pool of only \textit{a few thousand} static 3D scans (\textit{e.g.}, ScanNet). This reliance on a narrow set of environments fundamentally limits generalization, motivating us to construct a more scalable data curation framework. Instead of relying on a limited set of labeled 3D assets, we aim to unlock the potential of abundant web videos, automatically generating dense spatial annotations to fuel the next generation of spatial intelligence.

\subsection{Methods for Spatial Intelligence}
Efforts to enhance spatial intelligence in multimodal models primarily follow three approaches: (1) \textit{3D-native LMMs}, a line of works~\cite{deng20253d,zhu2024llava,wu2025spatial,mao2025spatiallm,zheng2025learning} directly consumes explicit 3D observations (\textit{e.g.}, point clouds, meshes, or multi-view RGB-D) and performs reasoning in 3D space, such as SpatialLM~\cite{mao2025spatiallm} and LLaVA-3D~\cite{zhu2024llava}. 
(2) \textit{2D-Centric spatial LMMs}~\cite{yang2025visual,cai2025scaling,yang2025cambrians,yin2025spatial,zhao2025spacemind} improve performance primarily by scaling up training datasets and training recipes on spatial perception and reasoning. For example, VST~\cite{yang2025visual} adopts 4.1M samples for SFT and RL, while Cambrian-S~\cite{yang2025cambrians} develops the VSI-590K dataset to enhance spatial video understanding. 
(3) \textit{3DGS-based methods}~\cite{qin2024langsplat,zou20253d,li2024langsurf} leverage 3DGS~\cite{kerbl20233d} as an explicit scene representation and optimize it to align geometry with language / vision signals~\cite{radford2021learning}. 
For example, M3-Spatial~\cite{zou20253d} augments a per-scene 3DGS reconstruction with language-aligned features to support open-vocabulary 3D grounding within the optimized scene. 

Despite notable progress, existing paradigms face fundamental scalability bottlenecks. Approaches (1) and (2) rely heavily on human-annotated 3D data or manually curated scans, making dataset expansion costly and intrinsically limited by 3D acquisition and annotation overhead. Approach (3), based on optimization-driven 3DGS, requires per-scene training or finetuning, which is time-consuming and often unstable, hindering large-scale deployment.
In contrast, Holi-Spatial reframes spatial data curation as a scalable, non-human pipeline that converts raw videos into \emph{annotated high-fidelity 3D scenes}, enabling automatic generation of large-scale spatial-understanding data.

\begin{figure*}
    \centering
    \includegraphics[width=1\textwidth]{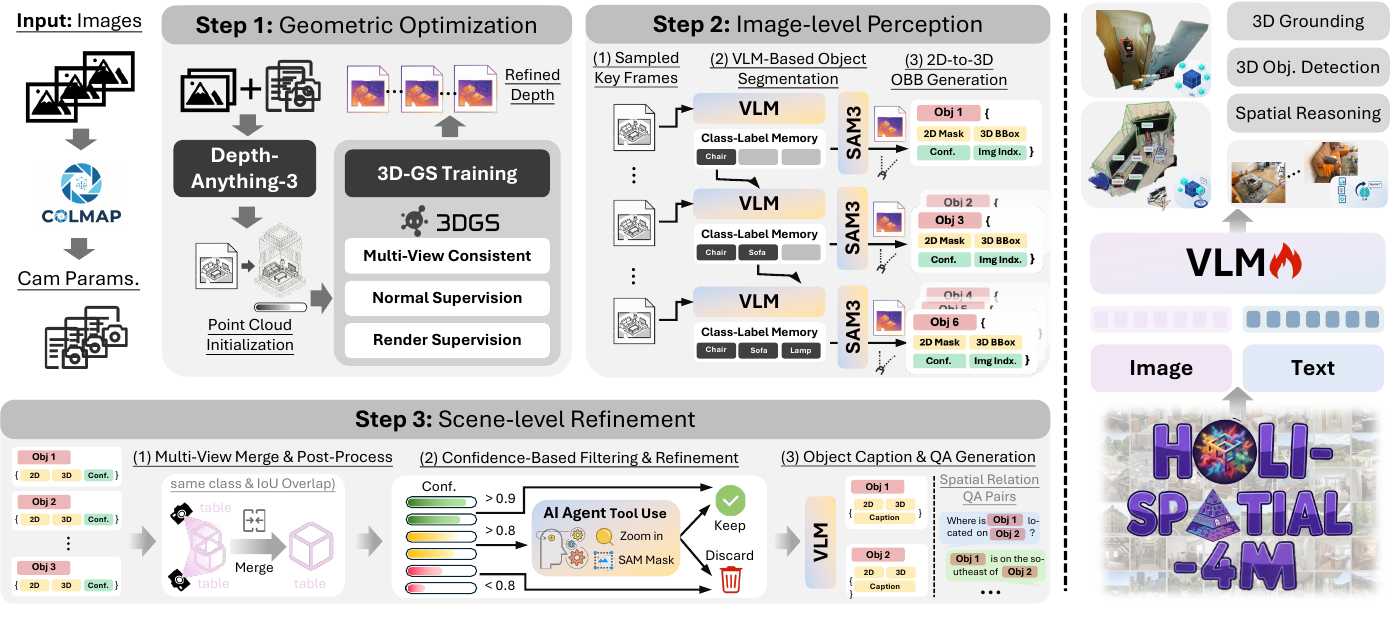}
    \vspace{-0.2in}
    \caption{\textbf{Overview of the \textbf{Holi-Spatial} data curation pipeline.} The framework operates in three progressive stages: (1) \textbf{Geometric Optimization} distills high-fidelity 3D structure from video streams using 3DGS; (2) \textbf{Image-level Perception} lifts 2D VLM and SAM3 predictions into initial 3D proposals; and (3) \textbf{Scene-level Refinement} employs a coarse-to-fine strategy to merge, verify, and caption instances, yielding dense, high-quality spatial annotations. Finally, leveraging the generated \textbf{Holi-Spatial-4M} dataset, we directly fine-tune the Qwen-VL family for downstream tasks (\textit{e.g.}, 3D grounding and spatial reasoning).}
    \label{fig:framework}
    \vspace{-0.2in}
\end{figure*}

\section{Method}
Here we introduce the core steps of our curation framework, as illustrated in~\cref{fig:framework}.
It is composed of three stages: (i) \textbf{Geometric Optimization} (\cref{sec:geo_opt}) distills high-fidelity 3D structure from raw videos; (ii) \textbf{Image-level Perception} (\cref{sec:img_percept}) extracts spatially consistent object labels from a VLM, produces high-quality 2D masks on keyframes, and lifts them into floor-aligned 3D proposals; and (iii) \textbf{Scene-level Refinement} (\cref{sec:scene_refine}) merges, filters, verifies, and captions instances in 3D to generate dense, reliable spatial annotations.

\subsection{Geometric Optimization}
\label{sec:geo_opt}
The primary objective of this stage is to distill raw video streams into high-fidelity geometric structures, serving as a robust prerequisite for precise spatial annotations. To achieve this, we first employ Structure-from-Motion~\cite{schoenberger2016sfm} to resolve accurate camera intrinsics and extrinsics, followed by leveraging a state-of-the-art spatial foundation model~\cite{lin2025depth} to initialize a dense, structurally coherent point cloud. However, feed-forward depth estimations inevitably contain noise and outliers. To address this, we incorporate 3D Gaussian Splatting (3DGS)~\cite{kerbl20233d} for per-scene optimization. Specifically, we integrate geometric regularization inspired by the surface reconstruction 3D Gaussian method~\cite{chen2024pgsr,li2024langsurf,zhangrade,Huang2DGS2024,gao2025citygs,Yu2024GOF} to enforce multi-view depth consistency. This process effectively eliminates large-scale floaters that would otherwise interfere with 3D bounding box generation.
These combined efforts yield a clean and consistent scene representation aligned with physical surfaces.

\begin{figure}[t]
    \centering
    \includegraphics[width=\linewidth]{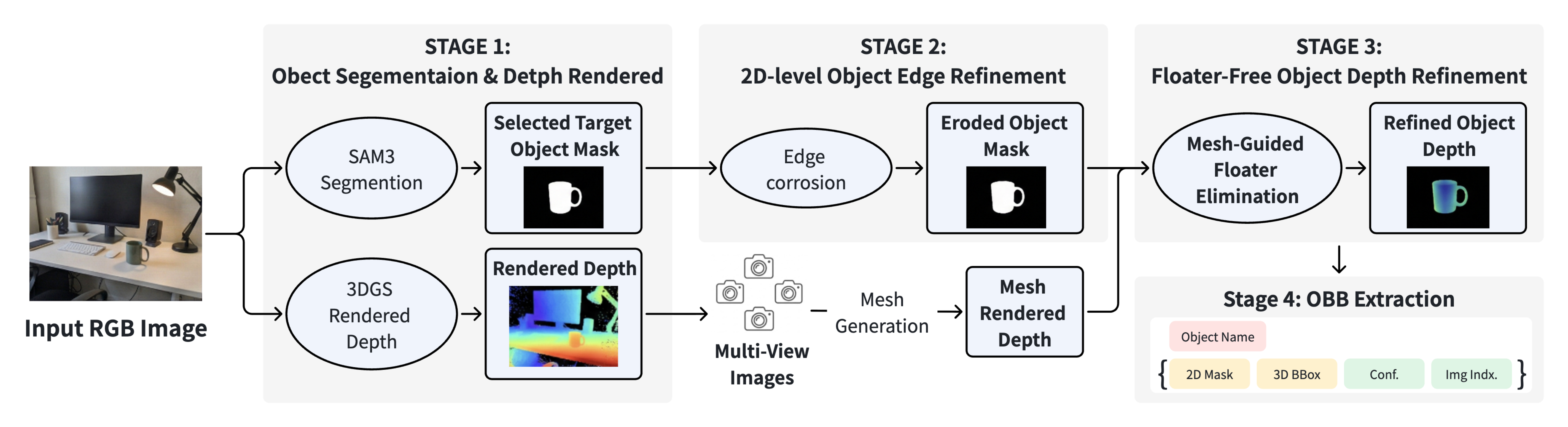}
    \caption{\textbf{Pipeline of 2D-to-3D OBB Generation.}
    We transform 2D object masks into initial 3D OBBs via depth projection, utilizing a four-step strategy to mitigate the impact of depth floaters.
    (1) We obtain an initial object depth map by combining 3DGS rendering with SAM3 instance segmentation.
    (2) To mitigate 2D boundary errors from SAM3, we erode the object mask near its contour and keep only the reliable interior region.
    (3) To remove 3D outliers caused by depth discontinuities, we use a multi-view-consistent mesh depth as guidance and filter inconsistent pixels in the 3DGS depth.
    (4) Finally, we estimate the initial 3D OBB from the refined point cloud, while preserving the associated 2D mask, confidence score, and source image index.
    }
    \label{fig:floter-masks}
\end{figure}

\subsection{Image-level Perception}
\label{sec:img_percept}
We uniformly sample a set of keyframes $\mathcal{I} = \{I_1, \dots, I_T\}$ from the raw video stream.
For each frame $I_t$, we employ Gemini3-Pro~\cite{team2023gemini} to generate a caption sequentially. 
Crucially, we maintain a dynamic class-label memory $\mathcal{M}_t$ to ensure semantic consistency. 
This memory accumulates recognized categories from previous frames ($I_{1:t-1}$) and instructs the VLM to prioritize reusing existing labels, formally updated as $\mathcal{M}_t = \mathcal{M}_{t-1} \cup \text{Extract}(I_t)$.
Guided by the prompt derived from $\mathcal{M}_t$, SAM3~\cite{carion2025sam} performs open-vocabulary instance segmentation, producing a set of predictions $\mathcal{O}_t = \{(M_k, s_k)\}_{k=1}^N$, where $M_k$ represents the binary mask and $s_k$ denotes the confidence score.

Leveraging the refined depth map $D_t$ rendered from our optimized 3DGS, we unproject each pixel $\mathbf{u}=(u,v)$ in mask $M_k$ into a 3D point $\mathbf{P} \in \mathbb{R}^3$ via:
$
    \mathbf{P} = D_t(\mathbf{u}) \cdot \mathbf{K}^{-1} \tilde{\mathbf{u}},
$
where $\mathbf{K}$ is the camera intrinsic matrix and $\tilde{\mathbf{u}} = [u, v, 1]^\top$ is the homogeneous coordinate. 
However, directly computing an Oriented Bounding Box (OBB) from these unprojected points often yields inaccurate bounds due to depth edge floaters around objects. 
These edge artifacts mainly arise from two sources: (i) \textbf{2D-level errors}, where SAM3 segmentation introduces boundary misalignment near object contours; and (ii) \textbf{3D-level noise}, where depth discontinuities naturally lead to unstable measurements and outliers. 
To suppress these floaters and improve grounding box precision, we seamlessly integrate a geometry-aware filtering strategy during the 2D-to-3D lifting process, as detailed in \cref{fig:floter-masks}.
Specifically, this strategy mitigates both the 2D boundary errors via mask erosion and the 3D outliers through mesh-guided depth filtering, ensuring that the estimated initial 3D OBBs are derived from a highly reliable geometry subset.

\begin{figure}[t]
    \centering
    \includegraphics[width=\linewidth]{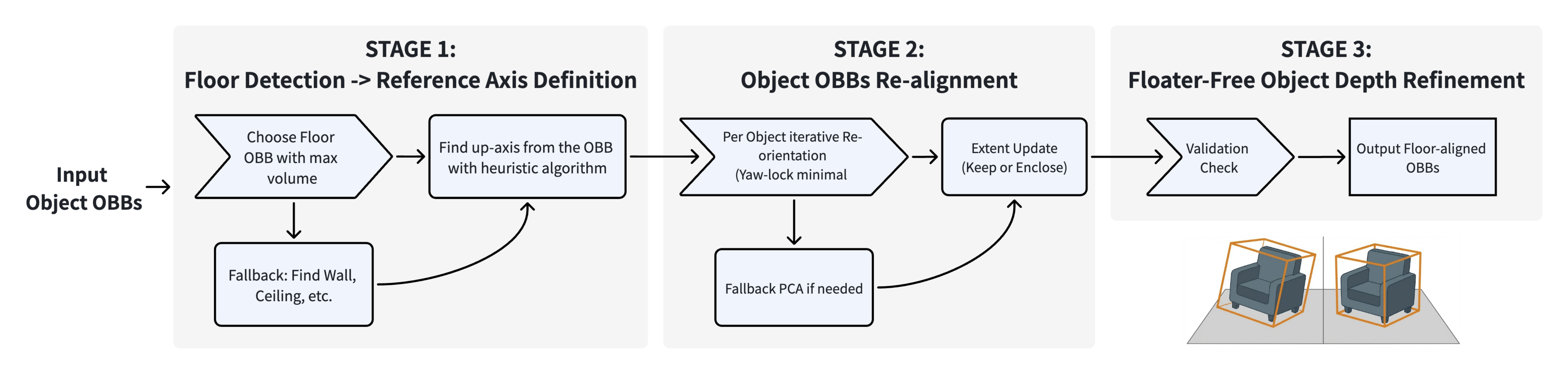}
    \caption{\textbf{Floor-aligned OBB post-processing pipeline.}
    Starting from the input instance OBBs produced after 2D-to-3D lifting and initial OBB estimation, we (1) detect a floor (or fallback planar structure) to infer a global up-axis, (2) re-align each instance OBB under a yaw-lock constraint with optional PCA fallback and update extents, and (3) apply a validation check to output gravity-/floor-consistent OBBs for downstream scene-level refinement.}
    \label{fig:floor_aligned_obb_pipeline}
\end{figure}

\subsection{Scene-level Refinement}
\label{sec:scene_refine}
The core motivation of this stage is a coarse-to-fine refinement strategy designed to distill high-fidelity annotations from noisy initial proposals. 
Based on the initial set of proposals $\mathcal{P}_{\text{init}} = \{(B_i, c_i, s_i)\}_{i=1}^M$ obtained from 2D-to-3D lifting, where $B_i \in \mathbb{R}^7$ denotes the raw 3D box parameters, $c_i$ is the semantic category, and $s_i$ is the confidence score, our refinement proceeds in three core parts:

\noindent
1) Multi-View Merge \& Post-Process. We first perform spatial clustering to consolidate redundant detections.
Specifically, we iterate over all pairs of instances $p_i, p_j \in \mathcal{P}_{\text{init}}$ and merge them if they share the same category and have sufficient 3D overlap:
\begin{equation}
    c_i = c_j \;\land\; \operatorname{IoU}_{3D}(B_i, B_j) > \tau_{\mathrm{merge}},
\end{equation}
where $\tau_{\mathrm{merge}}$ is set to $0.2$.
It mitigates object fragmentation by consolidating disjoint observations of the same object (\textit{e.g.}, parts of a large sofa) into a single 3D bounding box.
For each merged instance, we update its attributes to preserve the most reliable observation: the confidence score is set to $s_k = \max(s_i, s_j)$, and we retain the source image index associated with this maximum.
This selection mechanism grounds subsequent VLM captioning in the most informative visual perspective.
Following the merge, we apply a post-processing module to the resulting 3D OBBs to address any inconsistent roll and pitch caused by residual depth noise or partial observations. 
As detailed in \cref{fig:floor_aligned_obb_pipeline}, we perform a global gravity-alignment by estimating the global up-axis from the primary planar structure (\textit{e.g.}, floor) and re-orienting the vertical axis of each instance. 
This yields a fully consolidated, floor-aligned, and geometrically reliable set of instances, robustly prepared for downstream verification, which we denote as $\mathcal{P}_{\text{merged}}$.

\noindent
2) Confidence-Based Filtering \& Refinement. We then refine the consolidated set $\mathcal{P}_{\text{merged}}$ based on the updated scores $s_k$.
A tri-level decision rule is applied to each instance $p_k$:
\begin{equation}
\operatorname{Action}(p_k)=
\begin{cases}
\text{keep}, & s_k \ge \tau_{\mathrm{high}},\\
\text{discard}, & s_k < \tau_{\mathrm{low}},\\
\text{verify}, & \tau_{\mathrm{low}} \le s_k < \tau_{\mathrm{high}},
\end{cases}
\end{equation}
where thresholds are set to $\tau_{\mathrm{high}}=0.9$ and $\tau_{\mathrm{low}}=0.8$.
For proposals in the \textit{verify} band, we invoke a VLM-based agent equipped with an image zoom-in tool and a SAM3 re-segmentation tool to reassess the instance, yielding an updated confidence score $s'_k$.
We keep the proposal if $s'_k \ge \tau_{\mathrm{high}}$; otherwise, we discard it.

\noindent
3) Upon establishing the final set of validated instances $\mathcal{P}_{\text{final}}$, we proceed to generate dense semantic annotations. 
For each instance $p_k \in \mathcal{P}_{\text{final}}$, we retrieve the optimal source image $I_{k}^*$ corresponding to the retained high-confidence index from Step 1. 
Leveraging this most informative visual perspective, we employ Qwen3-VL-30B to generate a fine-grained caption for the object and procedurally synthesize a comprehensive suite of spatial QA pairs based on predefined templates, covering diverse tasks such as 3D grounding, spatial reasoning, and attribute identification.

\begin{figure*}[t]
    \centering
    \includegraphics[width=1\linewidth]{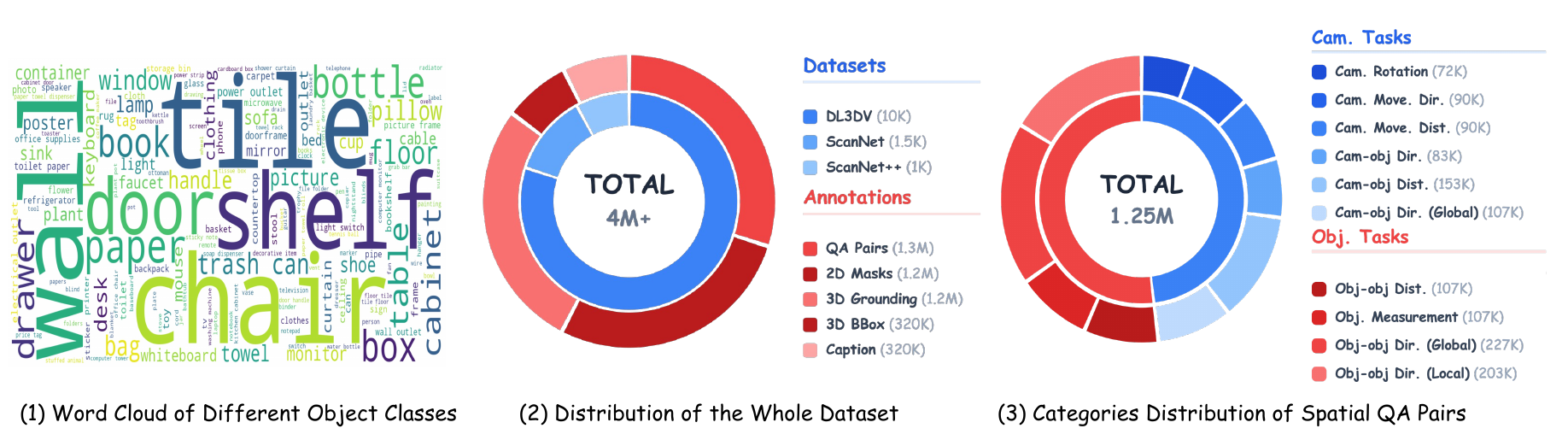}
  \caption{\textbf{Comprehensive Statistics of Holi-Spatial-4M.} (1) \textbf{Object Diversity:} Word cloud showing the long-tailed distribution of open-vocabulary categories. (2) \textbf{Dataset Composition:} The inner ring displays source scenes (ScanNet~\cite{dai2017scannet}, ScanNet++~\cite{yeshwanth2023scannet++}, DL3DV~\cite{ling2024dl3dv}), while the outer ring details the breakdown of over 4M generated spatial annotations. (3) \textbf{Spatial QA Taxonomy:} Distribution of 1.25M spatial QA pairs, categorized into Camera-centric tasks (\textit{e.g.}, rotation, movement) and Object-centric tasks (\textit{e.g.}, distance, direction).}
    \label{fig:dataset_stats}
    \vspace{-0.2in}
\end{figure*}
\section{Dataset}
We introduce \textbf{Holi-Spatial-4M}, the first large-scale, multi-spatial-modal dataset constructed using the proposed automated curation pipeline. As illustrated in~\cref{fig:dataset_stats}, this dataset represents a significant advancement in scale, granularity, and task diversity compared to existing datasets.

\noindent\textbf{Data Composition and Scale.}
Holi-Spatial-4M is derived from a diverse collection of raw video streams sourced from ScanNet~\citep{dai2017scannet}, ScanNet++~\citep{yeshwanth2023scannet++}, and DL3DV-10K~\citep{ling2024dl3dv}.
By processing these streams through our geometric optimization and scene-level refinement stages, we have curated over 12,000 optimized 3DGS scenes.
As shown in~\cref{fig:dataset_stats}~(2), the dataset encompasses a total of 4 million+ high-quality spatial annotations. This includes 1.3M 2D instance masks, 1.2M 3D grounding pairs, 320K 3D bounding boxes, and 320K detailed instance captions, significantly surpassing the scale of manual annotations in the original datasets.

\noindent\textbf{Open-Vocabulary Diversity.}
Unlike traditional datasets limited to a closed set of categories, Holi-Spatial-4M leverages the open-world knowledge of VLMs to annotate a vast array of objects. \cref{fig:dataset_stats}~(1) presents a word cloud of the object categories, highlighting the dataset's coverage of fine-grained indoor items. This semantic richness is crucial for training models capable of generalized spatial understanding in real-world environments.

\noindent\textbf{Spatial Question-Answering Pairs.}
To empower Large Multimodal  Language Models with robust spatial reasoning capabilities, we generate 1.25M Spatial QA pairs, structured into a comprehensive taxonomy.
As detailed in~\cref{fig:dataset_stats}~(3), these QAs are divided into two primary categories:
1) \textbf{Camera-centric Tasks} (blue sector), which challenge the model to understand ego-centric spatial changes such as \textit{Camera Rotation} and \textit{Movement Direction}; 2) \textbf{Object-centric Tasks} (red sector), which focus on allocentric reasoning, including \textit{Object-to-Object Distance}, \textit{Global/Local Direction}, and \textit{Size Measurement}.
This balanced distribution ensures that models trained on Holi-Spatial-4M develop a holistic understanding of 3D space.

\begin{table*}[t]
  \centering
  \caption{\textbf{Quantitative results of 3D spatial understanding on ScanNet, ScanNet++ and DL3DV datasets.} 
  We group methods by their input modality and task definition. 
  \textbf{3D Det}: 3D Object Detection (AP@25, AP@50); 
  \textbf{2D Seg}: 2D Object Segmentation (IoU); 
  \textbf{Depth}: Depth Estimation (F1-score). 
  \textbf{Bold} indicates best results. $\uparrow$: higher is better. \textcolor{gray}{`\textemdash'} denotes that the corresponding metric is not available. }
  \label{tab:spatial_understanding}
  \resizebox{\textwidth}{!}{
  \begin{tabular}{l cccc c cccc c cccc}
      \toprule[1.5pt]
      \multirow{3}{*}{\textbf{Method}} & 
      \multicolumn{4}{c}{\textbf{\emph{ScanNet}}} & &
      \multicolumn{4}{c}{\textbf{\emph{ScanNet++}}} & &
      \multicolumn{4}{c}{\textbf{\emph{DL3DV}}} \\
      \cmidrule(lr){2-5} \cmidrule(lr){7-10} \cmidrule(lr){12-15}
      
      & Depth ($\uparrow$) & 2D Seg. ($\uparrow$) & \multicolumn{2}{c}{3D Det. ($\uparrow$)} &
      & Depth ($\uparrow$) & 2D Seg. ($\uparrow$) & \multicolumn{2}{c}{3D Det. ($\uparrow$)} &
      & Depth ($\uparrow$) & 2D Seg. ($\uparrow$) & \multicolumn{2}{c}{3D Det. ($\uparrow$)} \\
      \cmidrule(lr){2-2} \cmidrule(lr){3-3} \cmidrule(lr){4-5}
      \cmidrule(lr){7-7} \cmidrule(lr){8-8} \cmidrule(lr){9-10}
      \cmidrule(lr){12-12} \cmidrule(lr){13-13} \cmidrule(lr){14-15}
      
      & F1 & IoU & AP$_{25}$ & AP$_{50}$ &
      & F1 & IoU & AP$_{25}$ & AP$_{50}$ &
      & F1 & IoU & AP$_{25}$ & AP$_{50}$ \\
      \midrule
      
      \multicolumn{15}{l}{\textit{\textbf{2D-VLM Methods}}} \\
      \hspace{3mm}SAM3 & \NA & 0.63 & \NA & \NA & & \NA & 0.50 & \NA & \NA & & \NA & 0.66 & \NA & \NA \\
      \hspace{3mm}SA2VA & \NA & 0.64 & \NA & \NA & & \NA & 0.25 & \NA & \NA & & \NA & 0.44 & \NA & \NA \\
      \addlinespace[3pt] 
      
      \multicolumn{15}{l}{\textit{\textbf{3D-VLM Methods}}} \\
      \hspace{3mm}SpatialLM & \NA & \NA & 11.42 & 8.19 & & \NA & \NA & 9.11 & 6.23 & & \NA & \NA & 7.05 & 4.38 \\
      \hspace{3mm}LLaVA-3D & \NA & \NA & 9.13 & 6.86 & & \NA & \NA & 12.2 & 4.80 & & \NA & \NA & 6.83 & 4.11 \\
      \hspace{3mm}SceneScript & \NA & \NA & 8.97 & 3.54 & & \NA & \NA & 9.86 & 4.42 & & \NA & \NA & 5.65 & 3.98 \\
      \addlinespace[3pt]
      
      \multicolumn{15}{l}{\textit{\textbf{3DGS-based Methods}}} \\
      \hspace{3mm}M3-Spatial & 0.32 & 0.22 & \NA & \NA & & 0.39 & 0.11 & \NA & \NA & & 0.23 & 0.13 & \NA & \NA \\
      \hspace{3mm}LangSplat & 0.19 & 0.36 & \NA & \NA & & 0.21 & 0.06 & \NA & \NA & & 0.18 & 0.24 & \NA & \NA \\
      
      \midrule
      \rowcolor{highlightgray}
      \textbf{Holi-Spatial (Ours)} & \textbf{0.98} & \textbf{0.66} & \textbf{76.60} & \textbf{67.00} & & \textbf{0.89} & \textbf{0.64} & \textbf{81.06} & \textbf{70.05} & & \textbf{0.78} & \textbf{0.71} & \textbf{62.89} & \textbf{52.67} \\
      \bottomrule[1.5pt]
  \end{tabular}
  }
\end{table*}

\section{Experiment}

\subsection{Framework Evaluation}

\begin{figure*}[b]
  \centering
  \includegraphics[width=0.95\linewidth]{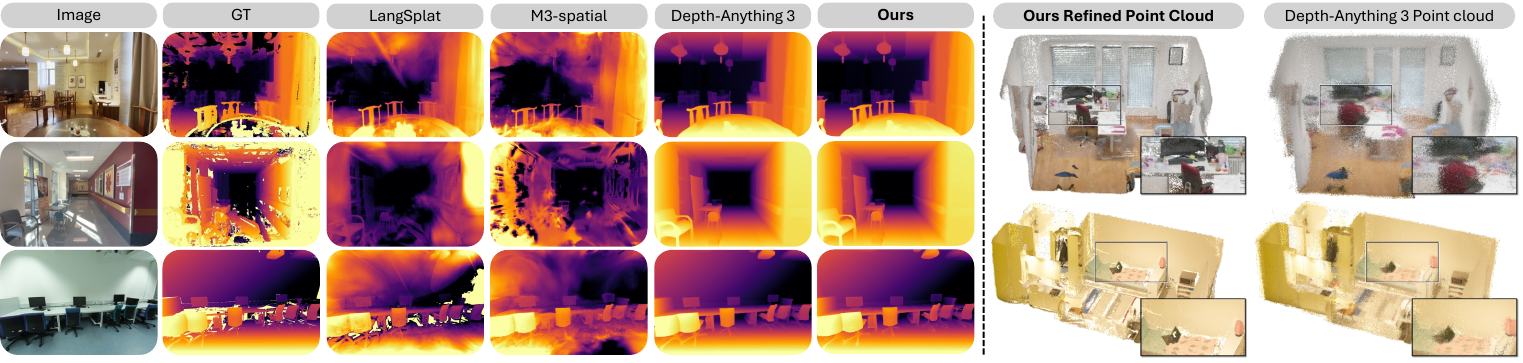}
  \caption{\textbf{Qualitative comparison of multi-view depth on ScanNet++.} We visualize depth maps from GT and baselines (LangSplat, M3-Spatial, Depth-Anything-V3) versus Holi-Spatial. Right: point clouds obtained by multi-view back-projection, where Holi-Spatial produces cleaner geometry with substantially fewer ghosting artifacts and floaters.}
  \label{fig:depth_vis}
      \vspace{-0.2in}
\end{figure*}

\noindent \textbf{Settings.} To conduct a fair comparison of our framework, we randomly sampled 10 scenes from each of the ScanNet, ScanNet++, and DL3DV-10K datasets. For these selected scenes, we manually annotated 2D instance masks and 3D bounding boxes to serve as the evaluation ground truth. For depth evaluation, we directly utilized the official ground truth depth maps provided by the respective scenes.

\noindent \textbf{Results.} \cref{tab:spatial_understanding} highlights Holi-Spatial as the sole framework capable of simultaneously generating high-quality predictions across 3D object detection, 2D segmentation, and depth estimation, whereas prior works typically specialize in single modalities. 
First, in terms of geometric fidelity, our method significantly outperforms 3DGS-based baselines with a Depth F1-score of 0.89 on ScanNet++~\cite{yeshwanth2023scannet++} compared to 0.39 for M3-Spatial~\cite{zou20253d}. As shown in~\cref{fig:depth_vis}, the quantitative depth visualization indicates that both DA3 and our method already provide strong relative depth cues compared with other baselines; to further highlight our multi-view advantage, the right panel visualizes the point cloud obtained by multi-view projection, where our results exhibit almost no ghosting and significantly fewer floaters.

\begin{figure*}[!t]
  \centering
  \includegraphics[width=\textwidth]{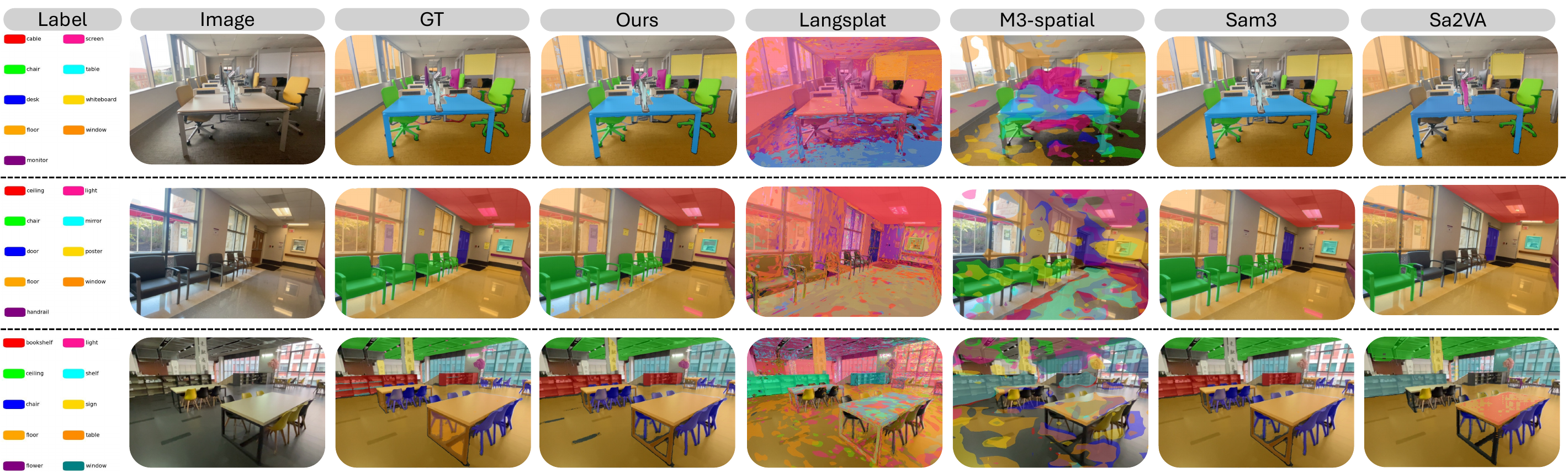}
  \caption{\textbf{Qualitative comparison of open-vocabulary 2D instance segmentation on the same scenes.} Holi-Spatial yields sharper boundaries, more complete masks under occlusion, and more accurate fine-grained categories.}
  \label{fig:2d_seg_vis}
\end{figure*}

\noindent
Regarding 2D segmentation, our framework demonstrates superior quality, reaching 0.64 IoU compared to 0.25 for SA2VA~\cite{yuan2025sa2va}. As illustrated in the second row of~\cref{fig:2d_seg_vis}, SAM3 fails to segment the distant mirror, whereas our method, by leveraging multi-view information, successfully segments such challenging and unclear instances. This improvement is attributed to the synergy between geometric and semantic priors: while the VLM agent provides robust semantic reasoning at the image level, our explicit integration of multi-view information compensates for incomplete observations from single images.

\begin{figure*}[!t]
  \centering
  \includegraphics[width=0.95\textwidth]{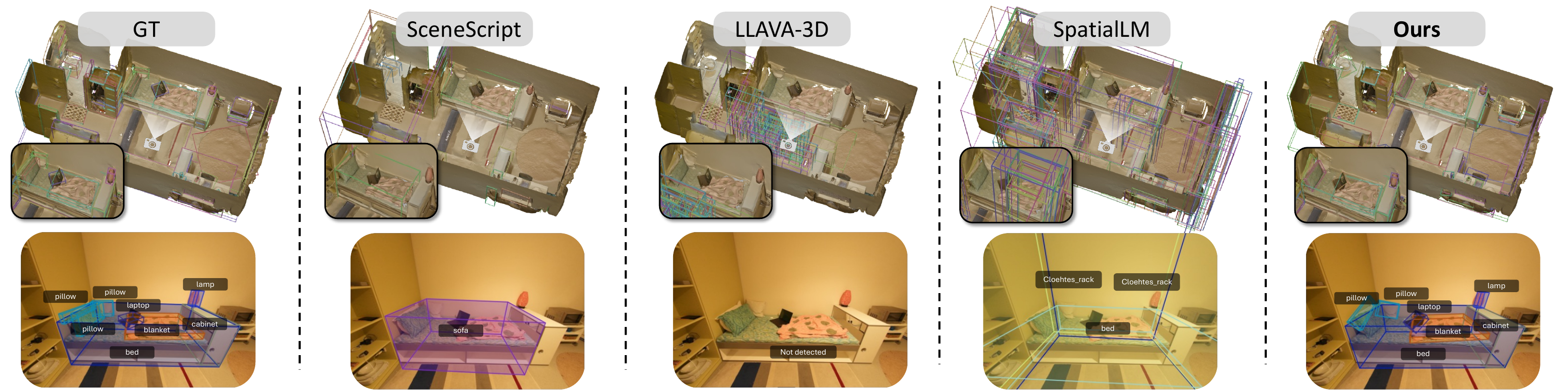}
  \caption{\textbf{Qualitative comparison of 3D object detection on ScanNet++.} We visualize predicted 3D bounding boxes from SceneScript, LLaVA-3D, SpatialLM, and Holi-Spatial. Holi-Spatial produces tighter boxes and more correct categories. }
  \label{fig:3d_detection}
      \vspace{-0.2in}
\end{figure*}

\noindent
Regarding 3D object detection, building on the high-quality geometry and refined semantics above, Holi-Spatial achieves dominant performance in 3D object detection. On ScanNet++, we report an AP$_{25}$ of 81.06, exceeding the state-of-the-art 3D-VLM (LLaVA-3D, 12.2 AP$_{25}$) by an order of magnitude. Furthermore, \cref{fig:3d_detection} provides a quantitative comparison of the predicted instances, where Holi-Spatial recovers substantially more objects with accurate labels and tightly aligned 3D bounding boxes, demonstrating strong semantic fidelity and geometric precision.

\subsection{VLM Finetuning Evaluation}

\begin{table}[!t]
  \centering
  \caption{\textbf{Quantitative results of Spatial Understanding QA tasks.} Here we caompare different models~\cite{ouyang2025spacer, zhu2025internvl3, wu2025spatial} on MMSI-bench~\cite{yang2025mmsi} and MindCube~\cite{yin2025spatial} benchmarks. \textbf{Bold} indicates the best performance.}
  \label{tab:model-comparison}
  \vspace{0.2cm} 
  \begin{tabular}{lcc}
    \toprule
    \textbf{Model} & \textbf{MMSI-Bench} & \textbf{MindCube} \\
    \midrule
    VST-SFT-3B~\cite{yang2025visual}              & 30.2 & 35.9 \\
    Cambrian-S-3B~\cite{yang2025cambrians}           & 25.2 & 32.5 \\
    VST-SFT-7B~\cite{yang2025visual}              & 32.0 & 39.7 \\
    Cambrian-S-7B~\cite{yang2025cambrians}           & 25.8 & 39.6 \\
    SpaceR-SFT-7B~\cite{ouyang2025spacer}           & 27.4 & 37.9 \\
    Intern3-VL-8B~\cite{zhu2025internvl3}           & 28.0 & 41.5 \\
    Spatial-MLLM~\cite{wu2025spatial}            & 27.0 & 32.1 \\
    
    \midrule
    Qwen3-VL-2B~\cite{Qwen3-VL}             & 26.1 & 33.5 \\ 
    Qwen3-VL-2B + Ours      & \textbf{27.6} & \textbf{44.0}\\ 
    \midrule
    Qwen3-VL-8B~\cite{Qwen3-VL}             & 31.1 & 29.4 \\
    Qwen3-VL-8B + Ours      & \textbf{32.6} & \textbf{49.1} \\ 
    \bottomrule
  \end{tabular}
\end{table}
\noindent \textbf{Settings}. For the spatial reasoning task, we finetune Qwen3-VL~\cite{Qwen3-VL} families using the 1.2M spatial QA pairs in our \textbf{Holi-Spatial-4M} dataset for 1 epoch with batch size 1024, and evaluate its performance on MMSI-Bench~\cite{yang2025mmsi} and MindCube~\cite{yin2025spatial}.
For 3D grounding task, we finetune Qwen3-VL-8B~\cite{Qwen3-VL} using 1.2M 3D grounding pairs in our Holi-Spatial-4M dataset. 
All models are trained for a single epoch with a total batch size of 1024. Training is conducted on 32 NVIDIA H800 GPUs (80GB).

\noindent \textbf{Spatial Reasoning}. 
\cref{tab:model-comparison} shows the QA training results. Our model consistently improves both 2B and 8B models after finetuning, improving model's spatial understanding thanks to our high quality 3D curated data and QA pairs. More QA examples are provided in the Appendix.

\begin{figure}[htb]
  \centering
  \includegraphics[width=0.6\linewidth]{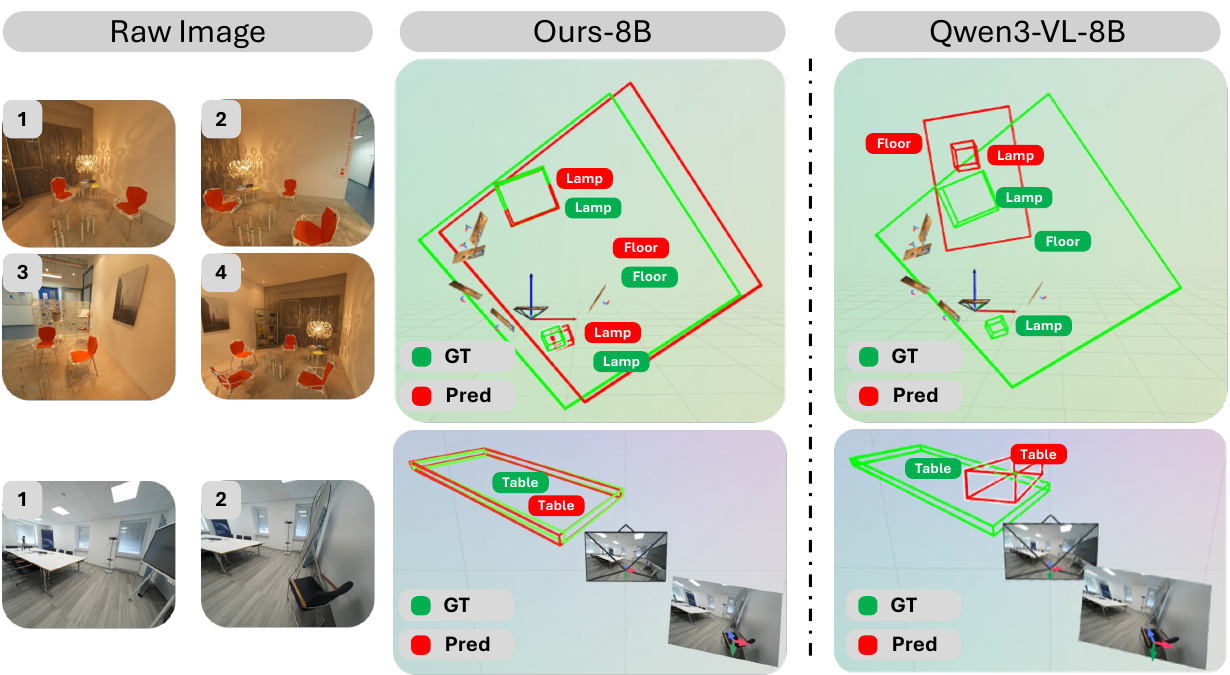}
  \caption{Given the same input images and queried objects, our predicted boxes align better with ground truth, indicating improved spatial localization.}
  \label{fig:vlm_grounding_vis}
\end{figure}

\noindent \textbf{3D Grounding}. The results are summarized in~\cref{tab:3d_grounding}. 
In particular, our method achieves an AP$_{50}$ of 27.98, exceeding the strongest baseline by 14.48 AP points, which we attribute to fine-tuning on our curated dataset with stronger 3D grounding supervision.
As illustrated in~\cref{fig:vlm_grounding_vis}, baseline models such as Qwen3-VL~\cite{Qwen3-VL}, trained primarily on single-view or anchor-view data, exhibit a clear viewpoint bias and fail to reliably ground objects across different views or at varying spatial depths.

\begin{table}[!t]
  \centering
  \footnotesize
  \caption{\textbf{Quantitative results of 3D Grounding on ScanNet++~\cite{yeshwanth2023scannet++} dataset.}}
  \label{tab:3d_grounding}
  \setlength{\tabcolsep}{6pt}
  \renewcommand{\arraystretch}{1.15}
  \begin{tabular}{lccc}
    \toprule
    \textbf{Method} & \textbf{AP$_{15}$} & \textbf{AP$_{25}$} & \textbf{AP$_{50}$} \\
    \midrule
    VST-7B-SFT~\cite{yang2025visual} & 17.29 & 14.50 & 11.20 \\
    Qwen3-VL-8B~\cite{Qwen3-VL} & 19.82 & 16.80 & 13.50 \\
    \midrule
    Qwen3-VL-8B + Ours & \textbf{35.52} & \textbf{31.94} & \textbf{27.98} \\
    \bottomrule
  \end{tabular}
      \vspace{-0.2in}
\end{table}

\begin{figure*}[htb]
  \centering
  \includegraphics[width=\linewidth]{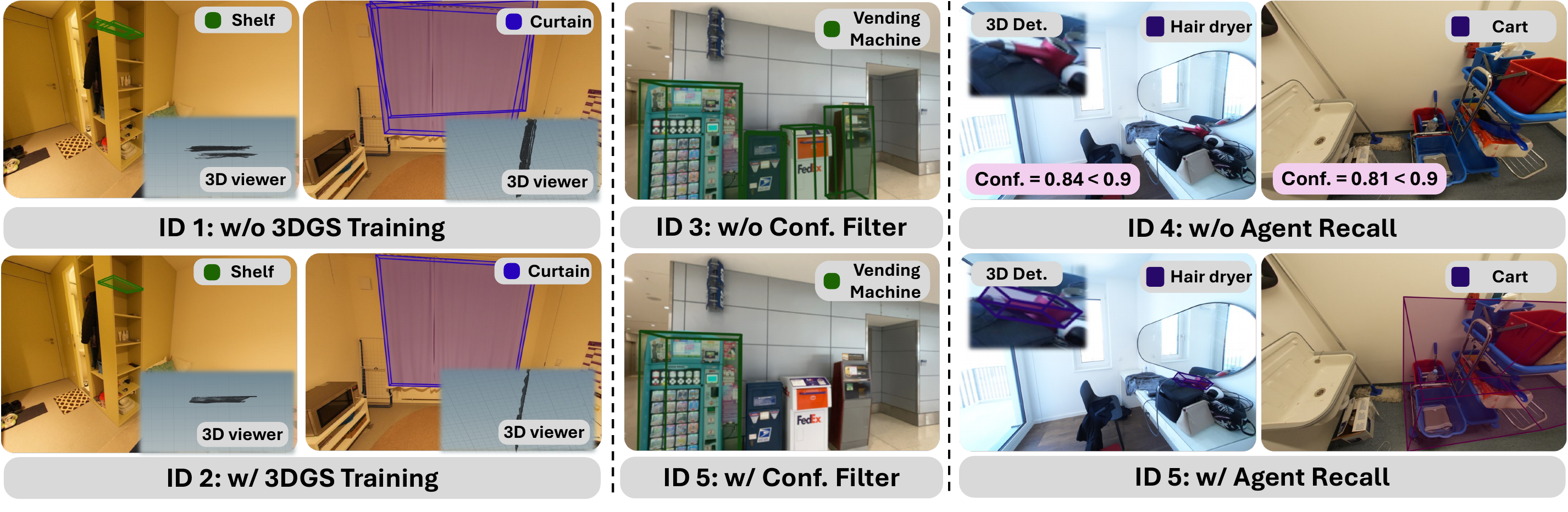}
  \caption{\textbf{Stage-wise visualization of scene-level refinement.} Detailed discussion is include in \Cref{sec:ablation}}
      \vspace{-0.2in}
  \label{fig:ablation_vis}
\end{figure*}

\subsection{Ablation}
\label{sec:ablation}

\noindent \textbf{Geometric Training}. In this section, we conduct ablation experiments on geometric training. As shown in~\cref{fig:ablation_vis}, we compare results obtained using the original DA3 depth (ID. 1) and those using our GS-refined depth (ID. 2) in the ScanNet++ scene. The visualization indicates that directly using DA3~\cite{lin2025depth} depth often introduces ghosting artifacts when projecting from multiple views. For example, in the shelf1curtain case, such ghosting leads to erroneous clustering and incorrectly splits one instance into multiple instances, which is also clearly observable in the 3D viewer at the bottom right. In contrast, our refined depth is more multi-view consistent, resulting in more reliable clustering and cleaner instance reconstruction. Moreover, as shown in~\cref{tab:merged_ablation}, the GS-refined depth substantially outperforms direct use of DA3~\cite{lin2025depth} depth.

\begin{table}[!t]
\centering
\caption{\textbf{Ablation study on depth refinement, confidence filtering, and agent recall.} ID. 1--2 (Step~1) compare depth sources, while ID. 3--5 (Step~3) evaluate post-processing modules. $P_{25}$ and $R_{25}$ denote precision and recall under IoU 25\%.}
\label{tab:merged_ablation}
\setlength{\tabcolsep}{5pt}
\renewcommand{\arraystretch}{1.15}
\sisetup{table-number-alignment = center}
\begin{tabular}{@{}c cccc S[table-format=2.2] S[table-format=2.2]@{}}
\toprule
\textbf{ID} & \makecell{\textbf{DA3}\\\textbf{Depth}} & \makecell{\textbf{3DGS}\\\textbf{Training}} & \makecell{\textbf{Conf.}\\\textbf{Filter}} & \makecell{\textbf{Agent}\\\textbf{Recall}} & {\textbf{$P_{25}$}} & {\textbf{$R_{25}$}} \\
\midrule

\rowcolor[HTML]{F2F2F2}
\multicolumn{7}{@{}l}{\textit{Step 1: Geometric Optimization}} \\
1 & \cmark & \xmark & \cmark & \cmark & 0.13 & 0.31 \\
2 & \cmark & \cmark & \cmark & \cmark & 0.81 & 0.89 \\
\addlinespace[2pt]

\rowcolor[HTML]{F2F2F2}
\multicolumn{7}{@{}l}{\textit{Step 3: Scene-Level Refinement}} \\
3 & \cmark & \cmark & \xmark & \xmark & 0.35 & 0.74 \\
4 & \cmark & \cmark & \cmark & \xmark & 0.67 & 0.69 \\
5 & \cmark & \cmark & \cmark & \cmark & 0.81 & 0.89 \\
\bottomrule
\end{tabular}
\end{table}

\noindent \textbf{Confidence Filter}.
As shown in~\cref{tab:merged_ablation}, we conduct ablation studies on the SAM3~\cite{carion2025sam} confidence filter mechanism in the ScanNet++ scene. The confidence filter improves precision (from 0.35 to 0.67) by suppressing false positives: as illustrated in~\cref{fig:ablation_vis} (ID. 3/ID. 4), SAM3 may assign an incorrect category (\textit{e.g.}, classifying multiple machines as \emph{vending machine}), and confidence-based filtering helps remove such misclassified predictions.

\noindent \textbf{Agent Refinement}.
According to our tests, as shown in~\cref{tab:merged_ablation}, Confidence filtering also reduces recall (from 0.74 to 0.69) when comparing ID. 3 and ID. 4, since it tends to discard visually challenging yet correct instances. For example, in~\cref{fig:ablation_vis}, ID. 4, it shows objects such as a hair dryer among clutter and a cart heavily occluded by buckets and cleaning tools can receive low confidence and thus be mistakenly filtered out.
To address this trade-off, we introduce an agent-based (VLM) verification step to reconsider borderline cases instead of discarding them directly, which recovers true positives with low confidence. Overall, the confidence filter and the agent recall complement each other and together yield the best precision--recall balance.

\section{Conclusion}
In this paper, we present \textbf{Holi-Spatial}, a fully automated pipeline that converts raw videos into high-fidelity 3D geometry and holistic spatial annotations. By combining 3DGS-based geometric optimization, open-vocabulary perception, and scene-level lifting/refinement, it produces multi-level supervision (rendered depth, 2D masks, 3D boxes, instance captions, and spatial QA). We further release \textbf{Holi-Spatial-4M} with 12K optimized 3DGS scenes, 1.3M masks, 320K 3D boxes/captions, 1.2M 3D grounding instances, and 1.2M QA pairs, and show on ScanNet/ScanNet++/DL3DV that fine-tuning VLMs on Holi-Spatial-4M consistently improves 3D grounding and spatial reasoning.

\section*{Impact Statement}
This work introduces Holi-Spatial, a fully automated data curation pipeline that converts raw video streams into high-fidelity 3D geometry together with holistic spatial annotations, enabling the construction of large-scale spatially-aware multimodal datasets. 

However, Holi-Spatial still has limitations. The pipeline relies on multiple upstream components and per-scene optimization, which can be computationally expensive and may degrade under challenging videos (e.g., limited viewpoints, motion blur, heavy occlusion, or dynamic objects). Open-vocabulary semantic labeling may also inherit biases or errors from foundation models, making robust verification and uncertainty estimation important future directions. We plan to improve efficiency (e.g., adaptive early stopping and better confidence-based validation), expand to broader domains and longer video contexts, and build stronger benchmarks for holistic 3D spatial understanding.

This work uses publicly available data sources and does not require collecting new sensitive personal data. Nonetheless, similar technology could be misused for privacy-invasive reconstruction of personal spaces; we encourage responsible deployment with consent, data governance, and appropriate safeguards.

\newpage

{
\small
\bibliographystyle{unsrt}
\bibliography{ref}
}

\appendix
\newpage
\renewcommand{\partname}{}
\part{Appendix} %
\parttoc %
\clearpage

\section{Why 3D Multi-view Merges}
\begin{figure}[htbp]
  \centering
  \begin{subfigure}[t]{0.48\linewidth}
    \centering
    \includegraphics[width=\linewidth]{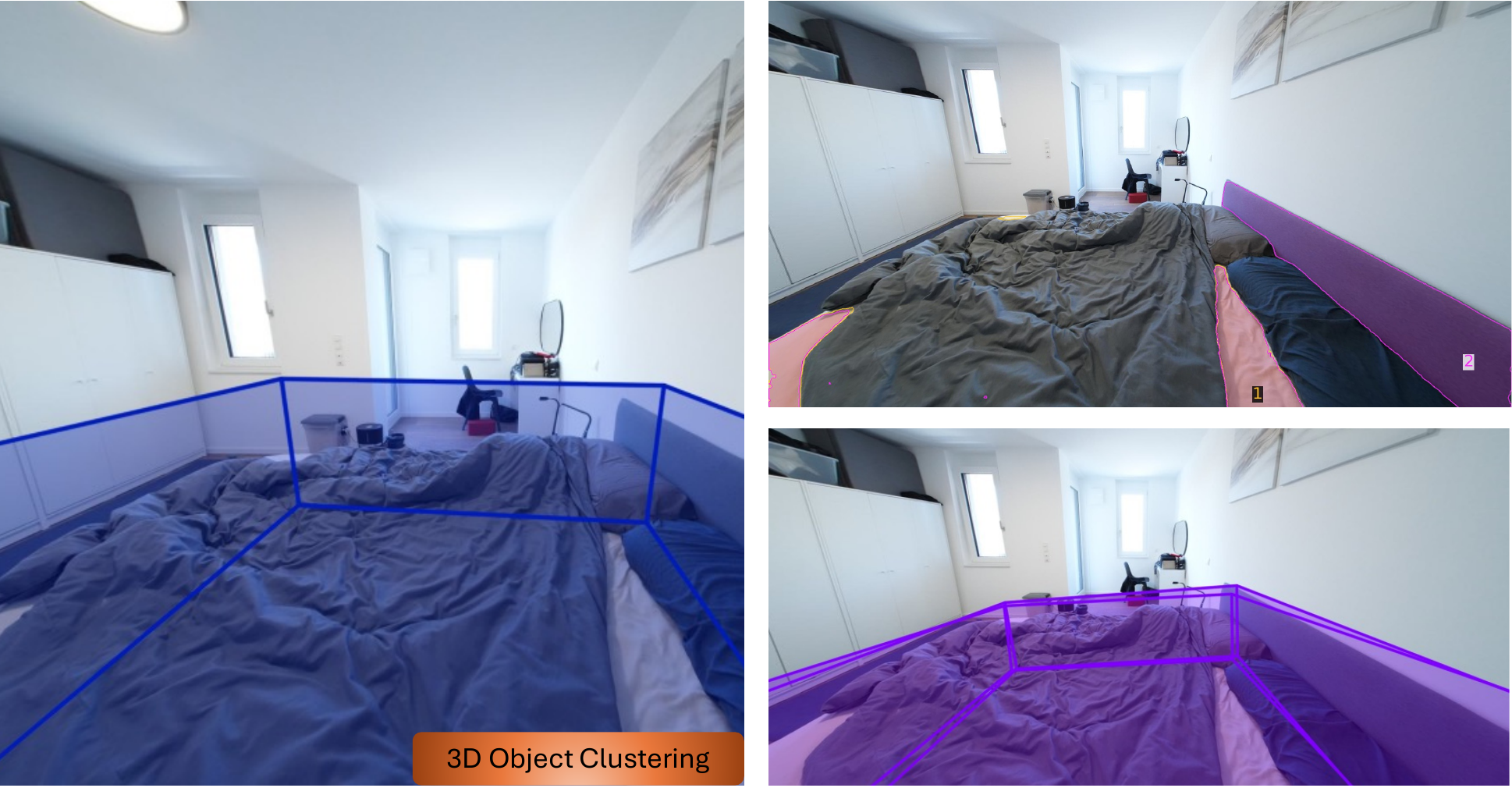}

    \label{fig:ab_3d_cluster}
  \end{subfigure}\hfill
  \begin{subfigure}[t]{0.48\linewidth}
    \centering
    \includegraphics[width=\linewidth]{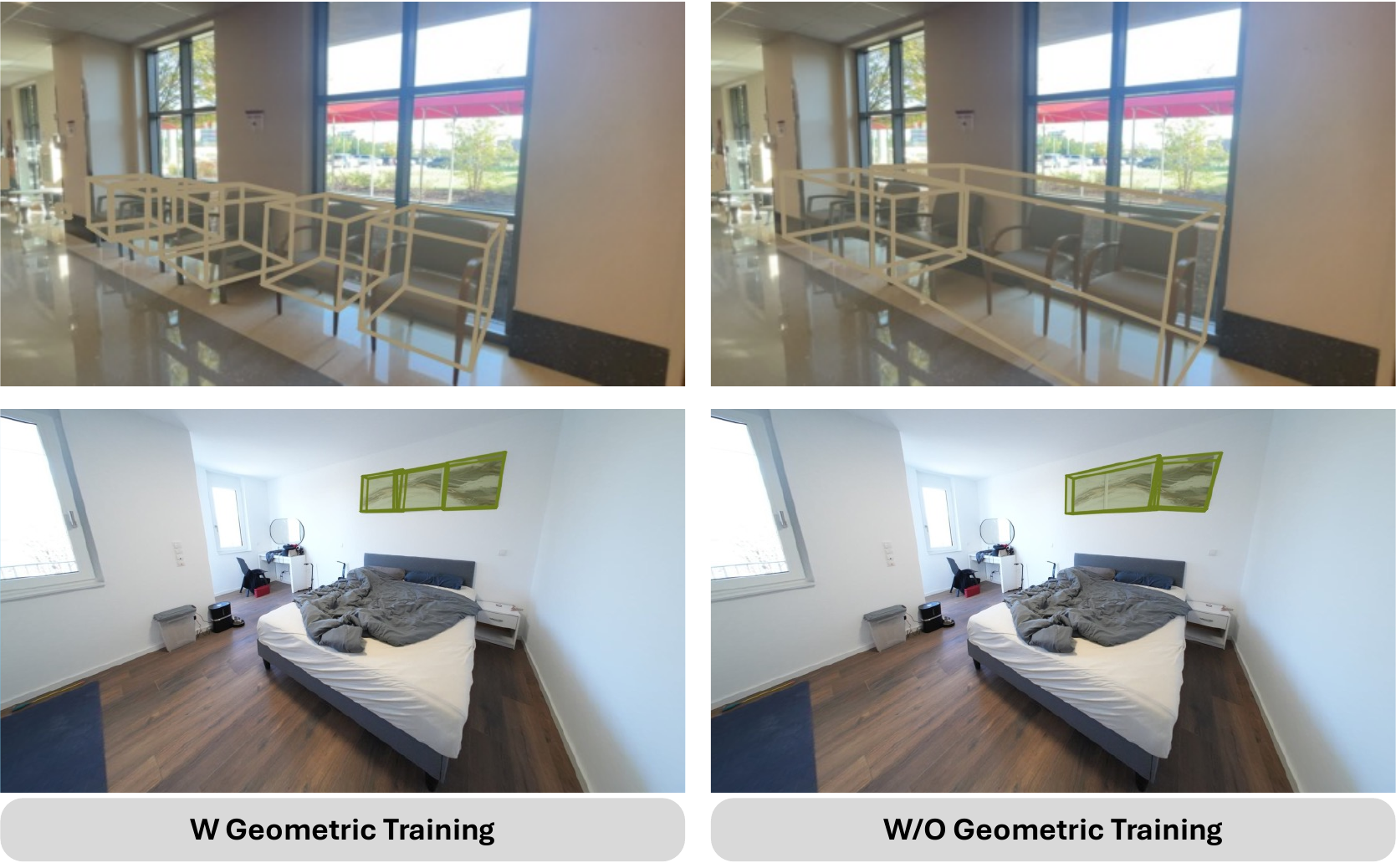}

    \label{fig:ab_da3_vs_gs_refi}
  \end{subfigure}
  \caption{\textbf{Ablation study on 3D multi-view merging and depth refinement.} (a) 3D geometric clustering corrects the fragmentation caused by occlusions in SAM3 image-level instance predictions, merging separated segments into unified object instances. (b) Using GS-refined depth constraints enables more accurate multi-view lifting, which avoids false merges and preserves the integrity of spatial object instances.}
  \label{fig:ab_da3_vs_gs_refined}
\end{figure}

Although the masks generated by SAM3 provide instance-level attributes at the image level, occlusions often cause SAM3 to fragment the same object into multiple instances---for example, in~\cref{fig:ab_3d_cluster}, a complete bed is detected as two separate beds. Thus, directly adopting SAM3's instance predictions is unreliable; robust 3D clustering and fusion remain essential.

For effective 3D clustering and merging, high geometric accuracy---namely, precise depth estimations---is crucial. As demonstrated in~\cref{fig:ab_da3_vs_gs_refined}, without the multi-view geometric constraint, estimated depth maps suffer from ghosting and aliasing artifacts, causing different objects to become spatially entangled and ultimately mis-merged as a single instance.

\begin{figure*}[!t]
  \centering
  \includegraphics[width=\textwidth]{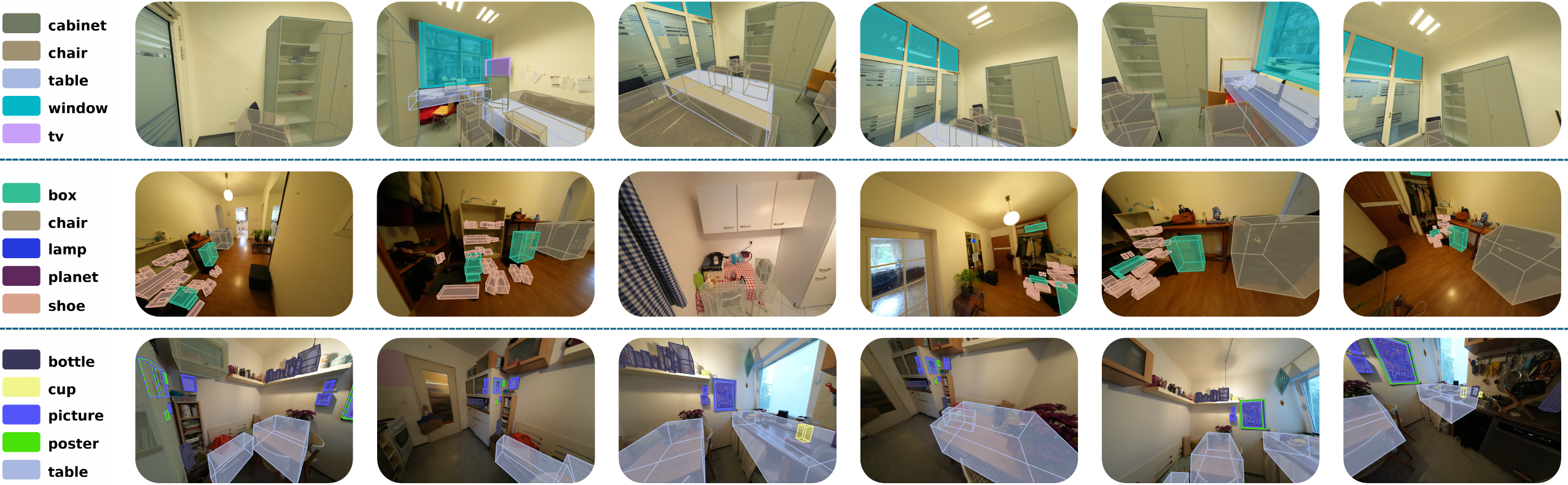}
  \caption{\textbf{Additional 3D detection visualizations.} We show Holi-Spatial predictions across diverse indoor scenes, demonstrating multi-view consistent localization and robust detection.}
  \label{fig:detection_box}
\end{figure*}

Furthermore, as shown in~\cref{fig:detection_box}, we provide more qualitative examples of 3D grounding box predictions across diverse object categories, underscoring the robustness and generalization capability of our method.

\begin{figure}[htbp]
  \centering
  \includegraphics[width=0.95\linewidth]{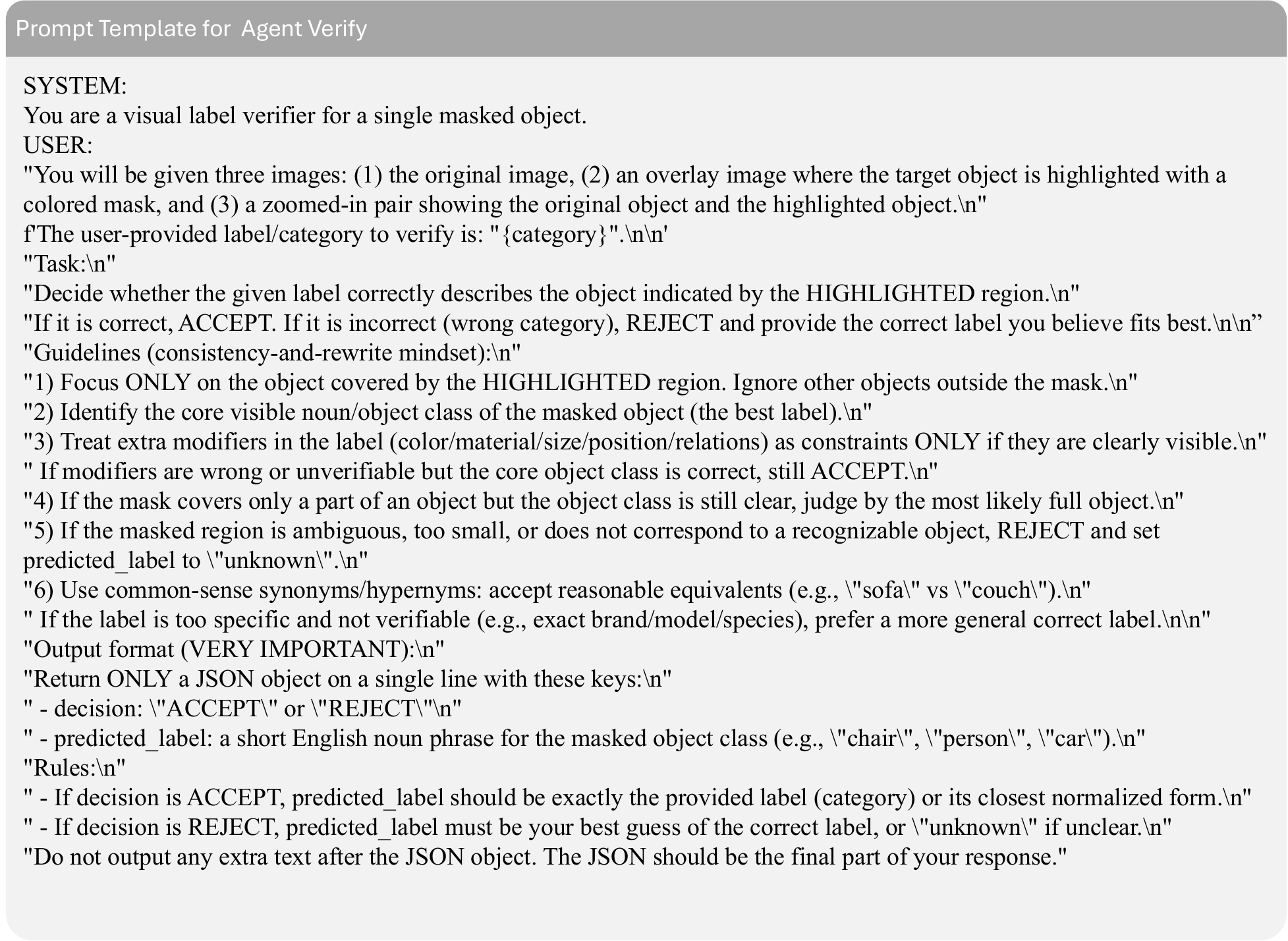}
  \caption{\textbf{Prompt template used for Agent verification.}}
  \label{fig:prompt_agent_verify}
\end{figure}

\begin{figure}[htbp]
  \centering
  \includegraphics[width=0.95\linewidth]{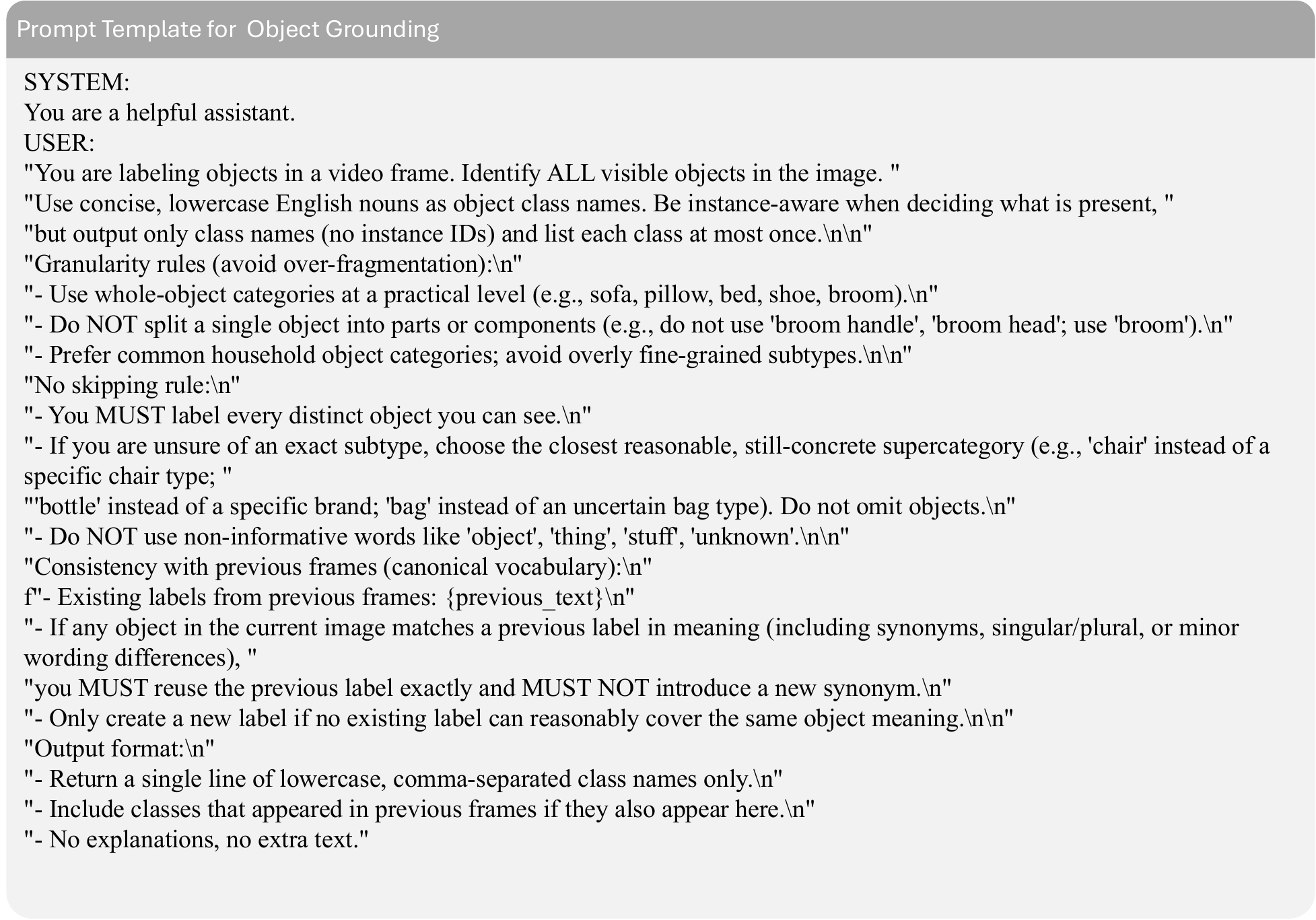}
  \caption{\textbf{Prompt template used for object grounding in 2D image.}}
  \label{fig:prompt_grounding_verify}
\end{figure}

\section{QA Examples}
\begin{figure}[t]
  \centering
  \includegraphics[width=0.95\linewidth]{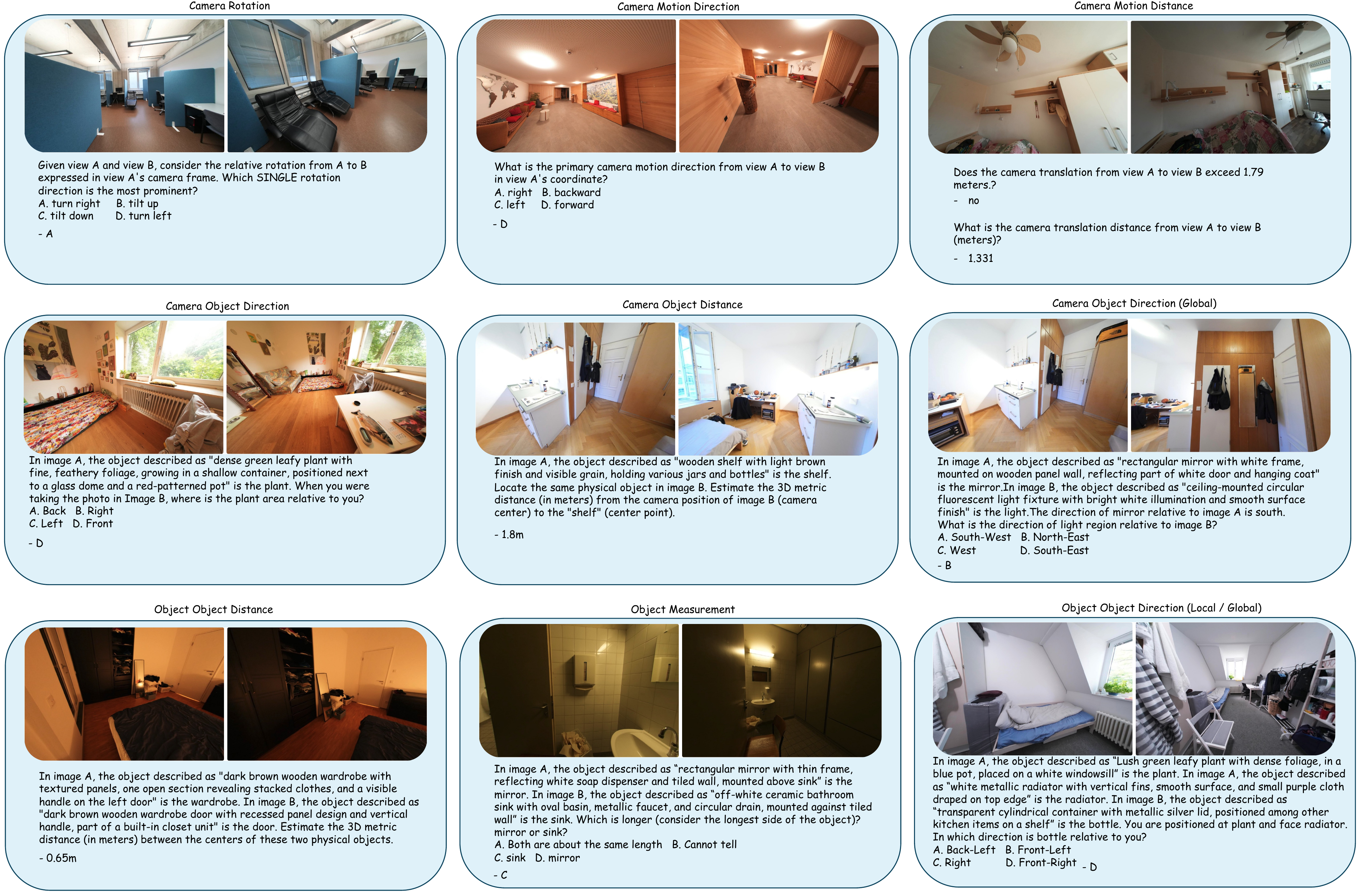}
  \caption{\textbf{Examples of 10 types of curated spatial QA pairs in Holi-Spatial.}}
  \label{fig:qa_examples}
\end{figure}

\begin{figure}[t]
  \centering
  \includegraphics[width=0.95\linewidth]{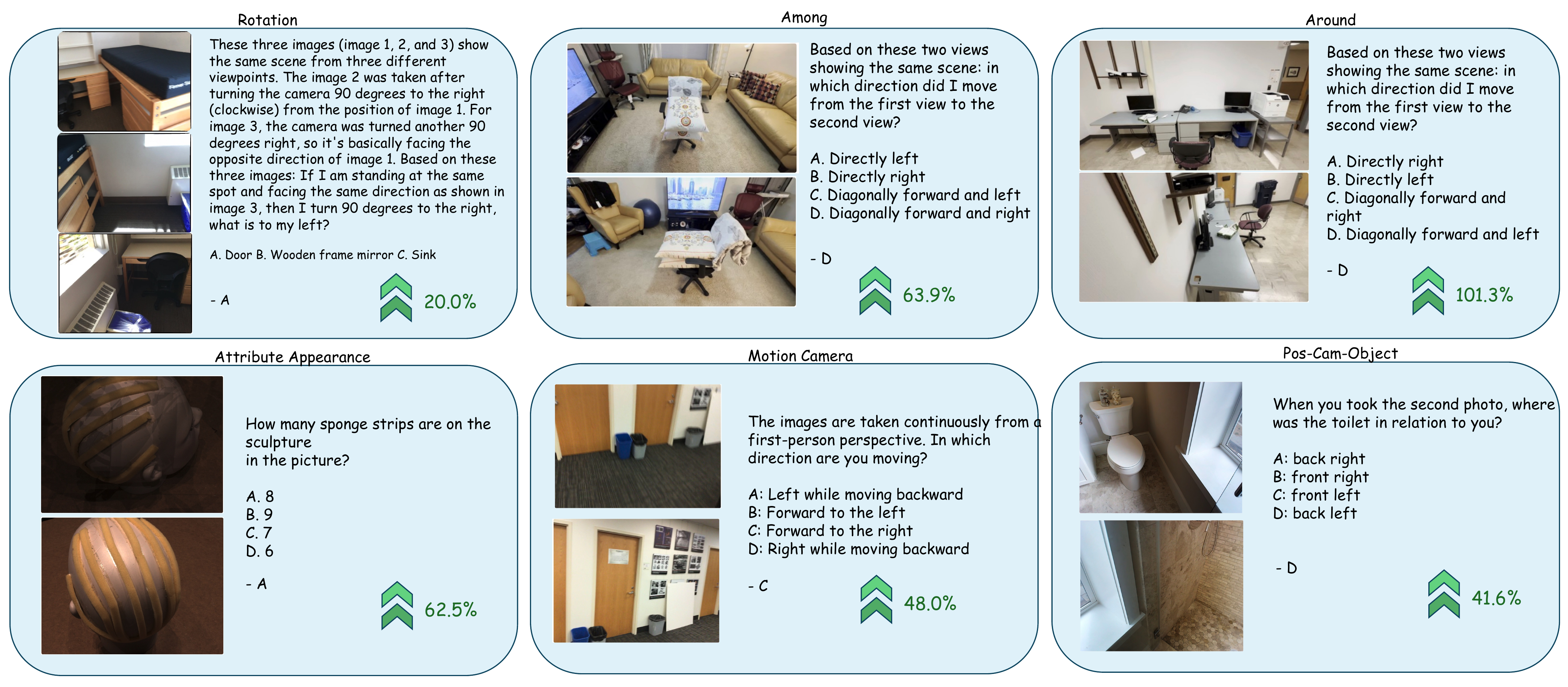}
  \caption{\textbf{Training with our curated QA data improves the three types of spatial reasoning tasks in MindCube~\cite{yin2025spatial} (upper row) as well as in MMSI-Bench~\cite{yang2025mmsi} (lower row).}}
  \label{fig:mindcube}
\end{figure}

We present examples of each of our curated QA question type in~\cref{fig:qa_examples}, including: camera rotation, camera movement direction, camera movement distance, camera-object direction, camera-object distance, camera-object distance (global frame), object-object distance, object measurement, object-object direction (local frame), object-object direction (global frame) . As shown in~\cref{fig:mindcube}, training on the QA dataset can greatly enhance many tasks in MindCube~\cite{yin2025spatial} and MMSI-Bench~\cite{yang2025mmsi}, especially perspective changing and egocentric imagination related questions. Additionally, we use VLM to describe the obejct's appearance, and use it as a way to reference the object in the question. This improves the appearance and grounding ability of models in spatial question answering, as shown in the bottom-left case in ~\cref{fig:mindcube}.

\section{Prompt Templates}
We provide the prompt templates for using agent verification in VLM-based confidence filtering and refinements in ~\cref{fig:prompt_agent_verify}, and the prompt templates for object grounding in 2D image in ~\cref{fig:prompt_grounding_verify}.

\section{Detail of Scene-Level Refinement Algorithm}

We provide the scene-level refinement algorithm in \cref{alg:scene_refinement}. 
The proposed scene-level refinement algorithm operates in three main stages to lift 2D open-vocabulary predictions into a globally consistent 3D representation. 

First, we project frame-level 2D masks generated by SAM3 into 3D space using depth maps and camera intrinsics, constructing localized point clouds and 3D bounding boxes for each semantic label. 

Next, to resolve cross-view redundancies, we spatially merge these candidates by clustering observations that share the same label and exhibit a 3D Intersection-over-Union (IoU) above a predefined threshold $\tau_{\mathrm{iou}}$. 

Finally, we employ a hybrid verification mechanism on the canonical view of each merged instance group: high-confidence instances ($s^\star \ge \tau_{\mathrm{high}}$) are directly accepted, low-confidence noise ($s^\star < \tau_{\mathrm{low}}$) is discarded, and ambiguous cases are rigorously arbitrated by a Vision-Language Model (VLM) using the cropped 2D canonical view. 

This pipeline effectively filters out spurious detections, yielding a robust and validated set of 3D instances $\mathcal{O}$ for the scene.

\begin{algorithm}[h]
\caption{\textbf{Scene-level Refinement via 3D Merging and VLM-based Verification}}
\label{alg:scene_refinement}
\begin{algorithmic}[1]
\Require keyframes $\{I_t\}_{t=1}^{T}$, camera parameters $\{\Pi_t\}_{t=1}^{T}$, refined depth maps $\{D_t\}_{t=1}^{T}$, per-frame label sets $\{\mathcal{L}_t\}_{t=1}^{T}$.
\Require $\mathrm{SAM3}(I_t,\ell)\rightarrow \{(m_{t,\ell,k}, s_{t,\ell,k})\}_{k=1}^{K_{t,\ell}}$, where $m$ is a 2D mask and $s\in[0,1]$ is confidence.
\Require thresholds $\tau_{\mathrm{iou}}$, $\tau_{\mathrm{low}} < \tau_{\mathrm{high}}$.
\Ensure validated instance set $\mathcal{O}$; each item $(\ell, B_g, t^\star, m^\star)$.

\vspace{0.25em}
\State \textbf{(1) Lift 2D instances to 3D candidates (per label)}
\State Initialize candidate pools $\mathcal{C}_\ell \gets \emptyset$ for all labels $\ell$.
\For{$t=1$ \textbf{to} $T$}
  \ForAll{$\ell \in \mathcal{L}_t$}
    \State $\{(m_{t,\ell,k}, s_{t,\ell,k})\}_{k=1}^{K_{t,\ell}} \gets \mathrm{SAM3}(I_t,\ell)$
    \For{$k=1$ \textbf{to} $K_{t,\ell}$}
      \State $P_{t,\ell,k} \gets \mathrm{BackProject}(m_{t,\ell,k}, D_t, \Pi_t)$
      \State $B_{t,\ell,k} \gets \mathrm{BBox}(P_{t,\ell,k})$
      \State $c_{t,\ell,k} \gets (P_{t,\ell,k}, B_{t,\ell,k}, s_{t,\ell,k}, t, m_{t,\ell,k})$
      \State $\mathcal{C}_\ell \gets \mathcal{C}_\ell \cup \{c_{t,\ell,k}\}$
    \EndFor
  \EndForAll
\EndFor

\vspace{0.25em}
\State \textbf{(2) Multi-view merging within each label (3D IoU)}
\State Initialize merged groups $\mathcal{G} \gets \emptyset$.
\ForAll{label $\ell$ such that $\mathcal{C}_\ell \neq \emptyset$}
  \State $\mathcal{G}_\ell \gets \mathrm{MergeByIoU3D}(\mathcal{C}_\ell, \tau_{\mathrm{iou}})$
  \ForAll{group $g \in \mathcal{G}_\ell$}
    \State $\mathcal{G} \gets \mathcal{G} \cup \{(\ell,g)\}$ \hfill \emph{// each $g$ is a set of candidates $c_{t,\ell,k}$}
  \EndForAll
\EndForAll

\vspace{0.25em}
\State \textbf{(3) Confidence gating and VLM-based verification}
\State Initialize $\mathcal{O} \gets \emptyset$.
\ForAll{instance group $(\ell,g) \in \mathcal{G}$}
  \State $c^\star \gets \arg\max_{c \in g}\; s(c)$ \hfill \emph{// canonical view}
  \State $(P^\star, B^\star, s^\star, t^\star, m^\star) \gets c^\star$
  \State $P_g \gets \bigcup_{c \in g} P(c)$
  \State $B_g \gets \mathrm{BBox}(P_g)$
  \If{$s^\star \ge \tau_{\mathrm{high}}$}
    \State $\mathcal{O} \gets \mathcal{O} \cup \{(\ell, B_g, t^\star, m^\star)\}$
  \ElsIf{$s^\star < \tau_{\mathrm{low}}$}
    \State \textbf{continue} \hfill \emph{// discard}
  \Else
    \State $y \gets \mathrm{VLMVerify}(I_{t^\star}, m^\star, \ell)$
    \If{$y=\texttt{true}$}
      \State $\mathcal{O} \gets \mathcal{O} \cup \{(\ell, B_g, t^\star, m^\star)\}$
    \EndIf
  \EndIf
\EndForAll

\State \Return $\mathcal{O}$
\end{algorithmic}
\end{algorithm}

\end{CJK*}
\end{document}